\documentclass{article}

\PassOptionsToPackage{numbers, compress}{natbib}

\usepackage[final]{neurips_2025}

\usepackage[utf8]{inputenc} 
\usepackage[T1]{fontenc}    
\usepackage{hyperref}       
\usepackage{url}            
\usepackage{booktabs}       
\usepackage{amsfonts}       
\usepackage{nicefrac}       
\usepackage{microtype}      

\usepackage{marvosym}
\usepackage{graphicx}
\usepackage{subcaption}
\usepackage{arydshln}
\usepackage{multirow}
\usepackage{xcolor}
\usepackage{colortbl}
\usepackage{wrapfig}

\usepackage{svg}
\usepackage{amsmath, amssymb, amsthm}
\usepackage{enumitem}
\usepackage{bm}
\usepackage[capitalize]{cleveref}
\definecolor{samsungblue}{RGB}{20, 40, 160}
\definecolor{samsungdark}{RGB}{10, 25, 80}

\hypersetup{
    colorlinks,
    linkcolor={samsungblue!80!black},
    citecolor={samsungblue},
    urlcolor={samsungdark}
}

\usepackage{color}
\usepackage{tabularx}
\usepackage{array}
\usepackage[export]{adjustbox}
\usepackage{booktabs}
\usepackage{titlesec}
\usepackage{xltabular}
\usepackage{siunitx} 
\usepackage{xspace}
\newcommand{\name}{\textsc{LittleBit}\xspace}

\newcommand*\boldell{\bm{\ell}}

\usepackage{soul}
\sethlcolor{yellow}

\newif\ifshowfixme 
\showfixmetrue

\ifshowfixme

  \newenvironment{tempblock}  
    {\begingroup\color{red}}  
    {\endgroup}  
\else

\fi  

\usepackage[dvipsnames]{xcolor}

\newtheorem{proposition}{Proposition}

\newtheorem{claim}{Claim}

\newcommand{\ie}{\textit{i}.\textit{e}., }
\newcommand{\eg}{\textit{e}.\textit{g}., }

\title{LittleBit: Ultra Low-Bit Quantization\\via Latent Factorization}

\author{%
  Banseok Lee$^{*}$ \quad Dongkyu Kim$^{*}$ \quad Youngcheon You \quad Youngmin Kim\textsuperscript{\dag}\\
Samsung Research \\
\texttt{\{bs93.lee, dongkyu.k, y01000.you, ym1012.kim\}@samsung.com}}

\makeatletter\def\Hy@Warning#1{}\makeatother

\begin{document}
\def\thefootnote{*}\footnotetext{Equal contribution}\def\thefootnote{\dag}\footnotetext{Corresponding author}

\maketitle

\begin{abstract}
The deployment of large language models (LLMs) is frequently hindered by prohibitive memory and computational requirements. While quantization mitigates these bottlenecks, maintaining model fidelity in the sub-1-bit regime remains a persistent challenge. In this paper, we introduce \name, a novel framework for extreme LLM compression. We target quantization rates as low as $0.1$ bits per weight (BPW), achieving a memory reduction of approximately $31\times$, which effectively compresses Llama2-13B to under $0.9$ GB. We represent weights via low-rank latent matrix factorization and subsequently binarize the resulting factors. To counteract the information loss inherent to such drastic precision reduction, we integrate a multi-scale compensation mechanism that learns importance parameters across row, column, and latent dimensions. Two primary contributions enable effective training: Dual Sign-Value-Independent Decomposition (Dual-SVID) for quantization-aware training (QAT) initialization, and Residual Compensation to minimize approximation errors. Extensive experiments confirm the superiority of \name in the sub-1-bit domain; for instance, our method at $0.1$ BPW surpasses the performance of leading techniques operating at $0.7$ BPW on Llama2-7B. We establish a new size-performance trade-off---unlocking a potential $11.6\times$ inference speedup relative to FP16---and render powerful LLMs practical for resource-constrained environments. Our code is available at \url{https://github.com/SamsungLabs/LittleBit}.
\end{abstract}

\section{Introduction}
\label{sec:introduction}

Large language models (LLMs) based on the Transformer architecture~\citep{attention2017} have fundamentally transformed natural language processing. However, the scale of these models, often reaching hundreds of billions or trillions of parameters~\citep{efficient2021, scaling2023}, results in prohibitive computational and memory costs. In particular, the high demand for GPU VRAM hinders widespread deployment on consumer or edge devices~\citep{edge2024, instinfer2024, dovetail2024}.

Model quantization, which reduces numerical precision, serves as a primary technique to address these bottlenecks~\citep{billm2024}. While post-training quantization (PTQ) methods such as GPTQ~\citep{gptq2023} and AWQ~\citep{awq2024} effectively compress models to approximately 4 bits, achieving further compression (\eg to 1-bit) typically requires quantization-aware training (QAT)~\citep{qat2018, qakd2019} to maintain performance. QAT has demonstrated the capability to enable 1-bit compression while preserving model fidelity~\citep{bitnet2023, onebit2024, fbillm2024}.

Nevertheless, even 1-bit models (\eg approximately 15.4 GB for 70B parameters~\citep{billm2024}) may exceed the capacity of highly resource-constrained devices~\citep{reviewEdgeLLM2024}. This limitation motivates the exploration of sub-1-bit quantization. Although prior work~\citep{stbllm2024} has investigated this area, maintaining performance at extremely low effective bits (\eg 0.55 bits per weight (BPW)) while ensuring computational efficiency without restrictive hardware dependencies remains challenging~\citep{stbllm2024}. Consequently, it is necessary to develop new techniques that achieve such extreme model compression, aiming for bit rates around 0.1 BPW, while preserving model performance.

\begin{figure}[t]
    \vspace{-10pt}
    \centering
    \includegraphics[width=0.90\linewidth]{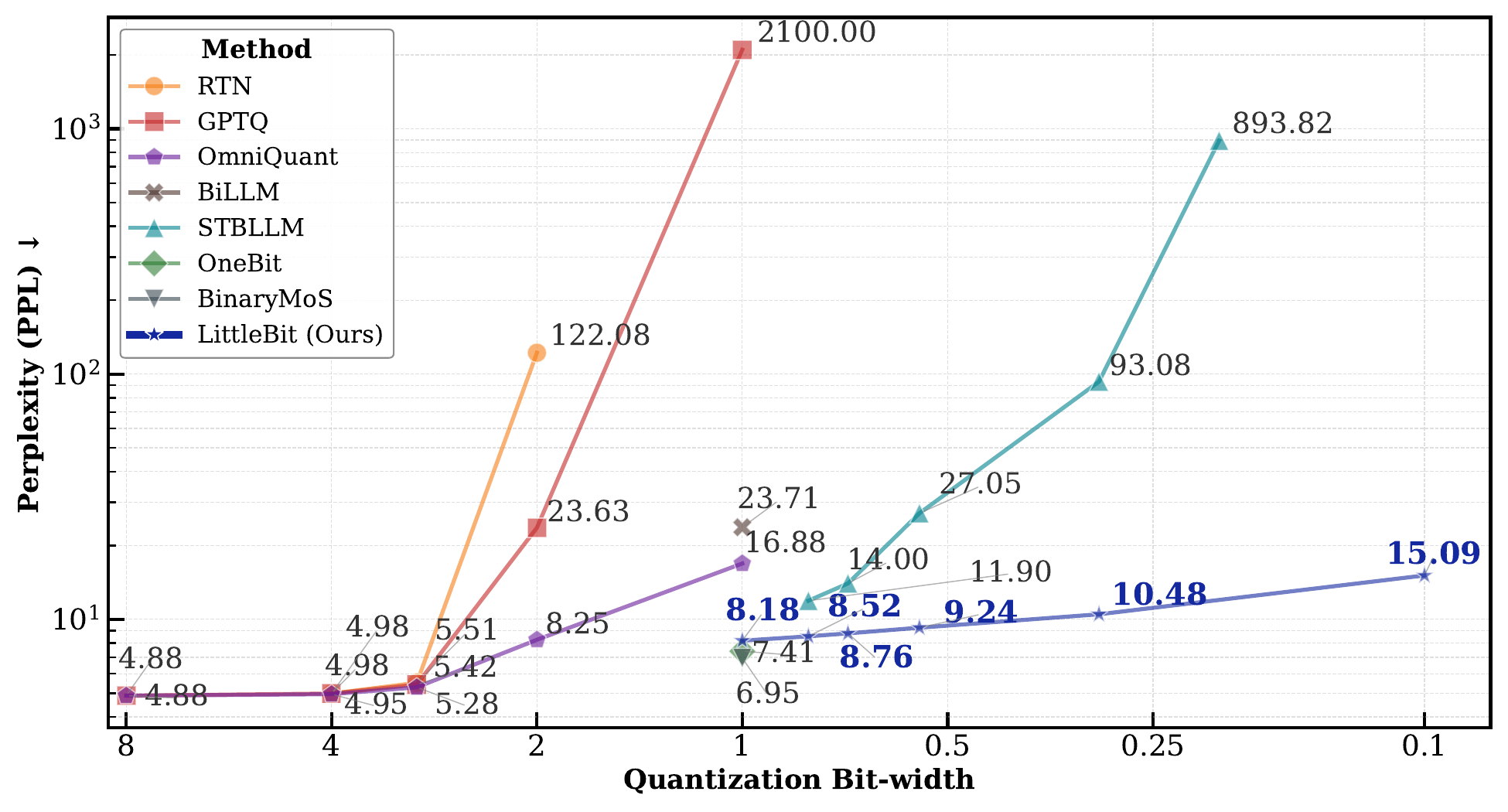}
    \caption{
      Low-bit quantization perplexity on Llama2-13B (WikiText-2). \name surpasses the state-of-the-art sub-1-bit quantization technique. Below 0.5 BPW, where the leading prior method degrades sharply, ours remains robust down to 0.1 BPW.
    } \label{fig:main_result}
    \vspace{-10pt}
\end{figure}

We derive our approach from two observations. First, LLM weight matrices often exhibit pronounced low-rank structure~\citep{galore2024, dobi2025}. This observation suggests that factorization methods such as Singular Value Decomposition (SVD)~\citep{svd1970} may offer a more stable compression pathway than pruning~\citep{deepcompression2016}, particularly under extreme compression regimes. For instance, at compression ratios exceeding 50\%, SVD-based approaches~\citep{svdllmv2_2025} consistently outperform pruning methods such as Wanda~\citep{wanda2023} (see \cref{appendix:pruning_vs_svd} for details). Second, binarization inherently causes information loss~\citep{survey2024}. Recent high-performing 1-bit methods~\citep{onebit2024,binarymos2024,arbllm2025} demonstrate that comprehensive scaling over multiple dimensions (\eg row and column) is crucial for mitigating this loss and stabilizing performance~\citep{arbllm2025}.

Building on these insights, we introduce \textbf{\name}, a novel method for extreme sub-1-bit quantization (\eg 0.1 BPW). In \name, we first represent weights via low-rank factorization ($\mathbf{W} \approx \mathbf{U}\mathbf{V}^\top$) and subsequently binarize these factors. To counteract errors from binarization, we leverage a multi-scale compensation mechanism that applies learnable scales across rows, columns, and an \textit{additional latent} dimension. We complement this architecture with Residual Compensation.

Extensive experiments demonstrate the superiority of \name over STBLLM~\citep{stbllm2024}, the leading sub-1-bit technique, on LLMs ranging from 1.3B to 32B parameters. Notably, \name achieves competitive performance on Llama2-13B at 0.1 BPW (\cref{fig:main_result}), surpassing STBLLM at 0.55 BPW. Furthermore, the 32B \name model at just 0.3 BPW maintains strong performance, substantially outperforming STBLLM and establishing a new state-of-the-art for sub-1-bit quantization.

In summary, our contributions are as follows:
\begin{itemize}[itemsep=0.0em]
    \item We propose \name, a novel framework unifying latent matrix factorization with multi-scale compensation to enable extreme sub-1-bit quantization. This method achieves effective bits down to 0.1 BPW while preserving model performance, establishing a viable size-performance trade-off for deploying LLMs in resource-constrained environments.
    \item We introduce Dual-SVID for the effective initialization of factorized structures and integrate Residual Compensation to mitigate approximation errors. These techniques collectively stabilize quantization-aware training (QAT) and counteract the information loss inherent in drastic precision reduction, ensuring robust learning dynamics.
    \item We provide extensive empirical validation demonstrating that \name consistently outperforms STBLLM~\citep{stbllm2024}, the prior leading sub-1-bit technique, across various model scales. Our results confirm that \name maintains superior fidelity in the deep compression regime, where baseline methods typically suffer severe degradation.
\end{itemize}

\section{Related Works}
\label{sec:related_works}

\subsection{Binarization and Quantization}
\label{sec:binarization}
Network binarization aims for extreme compression and acceleration by constraining weights or activations to $\pm 1$. This enables efficient bitwise operations~\citep{binarized2016, towards2018}. While early studies such as BinaryConnect~\citep{binaryconnect2015} and BNNs~\citep{binarized2016} demonstrated feasibility, they encountered substantial accuracy degradation~\citep{binarysurvey2020}. Researchers partially mitigated this degradation by introducing scaling factors~\citep{xnor2016} and employing the Straight-Through Estimator (STE)~\citep{ste2013} to handle the non-differentiable quantization function during training~\citep{custom2024}.

The direct application of these early techniques to large language models (LLMs) typically results in severe performance loss~\citep{pbllm2023}. Consequently, LLM-specific quantization strategies have emerged. These are primarily categorized into post-training quantization (PTQ) and quantization-aware training (QAT)~\citep{survey2021}. PTQ adapts pre-trained models and achieves competitive results at approximately 4-bit precision with methods such as GPTQ~\citep{gptq2023} and AWQ~\citep{awq2024}. However, these methods generally struggle to maintain performance below 2-bit precision~\citep{pbllm2023, billm2024}. The sub-1-bit regime presents particular challenges for PTQ methods due to severe information loss~\citep{predicting2013}. Recent work such as STBLLM~\citep{stbllm2024} demonstrates that incorporating structured binarized compression can extend PTQ into the sub-1-bit regime, but a non-negligible accuracy gap persists.

In contrast, QAT integrates the quantization process directly into the training loop. This enables the model to learn and adapt to the low-precision format~\citep{llmqat2023}. This approach generally yields superior performance compared to PTQ at very low bit-widths. Successful QAT approaches for low-bit scenarios often employ sophisticated scaling mechanisms alongside adaptive training, as evidenced by recent methods~\citep{onebit2024, binarymos2024}. Despite these advancements, consistently achieving high model fidelity when quantizing to less than $1.0$ BPW remains a considerable challenge. This persisting difficulty underscores the need for novel architectures and training strategies specifically designed for such extreme compression regimes.

\subsection{Low-Rank Approximation in Quantization}
\label{sec:low_rank}
Beyond quantization, the inherent parameter redundancy within deep neural networks~\citep{predicting2013} can be effectively exploited using low-rank approximation methods. This redundancy is particularly evident in LLMs~\citep{shortgpt2024,whatmatters2024}. Techniques such as Singular Value Decomposition (SVD)~\citep{svd1970} allow large weight matrices to be represented as products of smaller factors. This offers an alternative compression strategy that may be more resilient than pruning, especially under high compression regimes~\citep{deepcompression2016}. Recognizing the potential for complementary benefits, researchers have explored combining low-rank factorization to reduce parameter count with quantization to lower parameter precision~\citep{caldera2024}. Initial studies applied SVD in various roles within the LLM context. Some approaches used activation statistics to guide decomposition~\citep{svdllmv2_2025}, while others integrated SVD principles into initialization or parameter-efficient updates~\citep{adalora2023, pissa2024, onebit2024}. Further works applied transformations prior to factorization~\citep{asvd2023} or employed low-rank structures to model quantization errors~\citep{lqer2024}.
\begin{figure}[t]
    \vspace{-5pt}
    \centering
    \includegraphics[width=0.85\linewidth]{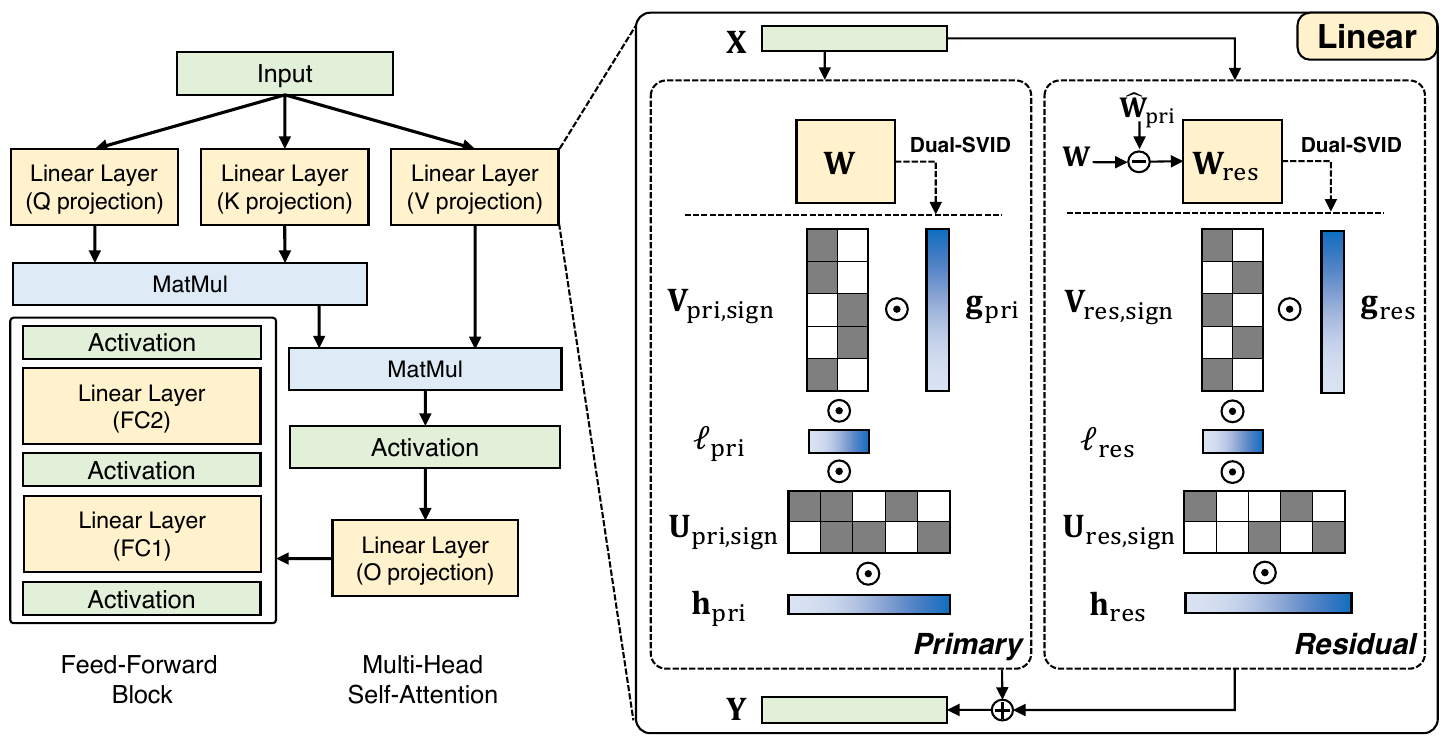}
    \caption{Comparison of a standard Transformer layer (left) and the \name architecture (right). \name performs linear transformation using parallel Primary and Residual pathways. The Primary path employs binarized factors ($\mathbf{U}_\mathrm{sign}, \mathbf{V}_\mathrm{sign}$) and FP16 scales ($\mathbf{h}, \mathbf{g}, \boldell$) on input $\mathbf{X}$, initialized from $\mathbf{W}$ via Dual-SVID. Simultaneously, the Residual path computes a correction with its own parameters ($\mathbf{U}_{\mathrm{res, sign}}, \mathbf{V}_{\mathrm{res, sign}}, \mathbf{h}_\mathrm{res}, \mathbf{g}_\mathrm{res}, \boldell_\mathrm{res}$) from the approximation residual. Their outputs sum to form $\mathbf{Y}$, eliminating storage of the effective weight matrices $\widehat{\mathbf{W}}_\mathrm{pri}$ and $\widehat{\mathbf{W}}_\mathrm{res}$.}
    \label{fig:littlebit_arch}
\end{figure}
A more integrated approach involves incorporating low-rank concepts directly within QAT methods. For example, DL-QAT~\citep{dlqat2024} jointly learns a representation amenable to both low-rank approximation and low-precision quantization. However, this method is currently limited to 3-bit quantization. Optimizing structural properties such as rank and numerical precision during training represents a promising path toward effective LLM compression. There is potential for further exploration of even lower precision quantization.

\section{Methodology}
\label{sec:methodology}
This section details \textbf{\name}. It describes the factorized architecture with multi-scale compensation (\cref{sec:little_bit_arch}), the Dual-SVID initialization for stable QAT (\cref{sec:dual_svid}), and Residual Compensation for enhanced fidelity (\cref{sec:residual_compensation}). These components are optimized via QAT, as detailed in \cref{sec:experiments}.

\subsection{\name Architecture}
\label{sec:little_bit_arch}
The \name architecture, depicted in \cref{fig:littlebit_arch}, redesigns linear layers for extreme compression (\eg 0.1 BPW) by synergistically combining low-rank factorization and binarization. The method leverages the observed low-rank structure in LLM weights~\citep{svdllmv1_2025} to approximate the weight matrix $\mathbf{W} \in \mathbb{R}^{d_\mathrm{out} \times d_\mathrm{in}}$ as:
\begin{equation}
\label{eq:svd}
\mathbf{W} \approx \mathbf{U} \mathbf{V}^\top \quad (\mathbf{U} \in \mathbb{R}^{d_\mathrm{out} \times r}, \mathbf{V} \in \mathbb{R}^{d_\mathrm{in} \times r}, r \ll \min(d_\mathrm{in}, d_\mathrm{out}))
\end{equation}
We prioritize this factorization-based approach over other structural compression methods, such as pruning, due to its superior robustness in extreme compression regimes. As the comparative analysis against methods like Wanda demonstrates (\cref{appendix:pruning_vs_svd}), SVD-based compression maintains model integrity far more effectively at high compression ratios. This provides a more stable foundation for ultra low-bit quantization.
The resulting factors $\mathbf{U}$ and $\mathbf{V}$ are then binarized:
\begin{equation}
\label{eq:uv_sign}
\mathbf{U}_\mathrm{sign} = \mathrm{sign}(\mathbf{U}), \quad \mathbf{V}_\mathrm{sign} = \mathrm{sign}(\mathbf{V}) \in \{-1, +1\}
\end{equation}
To compensate for the significant information loss incurred by binarization~\citep{survey2024}, \name incorporates learnable FP16 scales. Beyond standard row ($\mathbf{h}$) and column ($\mathbf{g}$) scaling~\citep{pbllm2023,onebit2024,arbllm2025}, the architecture introduces an \textit{additional latent} scale ($\boldell \in \mathbb{R}^r$). This latent scale addresses the factorization structure by learning the relative importance of each of the $r$ latent dimensions, which correspond to the columns of the binarized factors $\mathbf{U}_\mathrm{sign}$ and $\mathbf{V}_\mathrm{sign}$:
\begin{equation}
\label{eq:ghl_scales_combined}
\mathbf{h} \in \mathbb{R}^{d_\mathrm{out}}, \quad \mathbf{g} \in \mathbb{R}^{d_\mathrm{in}}, \quad \boldell \in \mathbb{R}^{r}
\end{equation}
The primary effective weight, $\widehat{\mathbf{W}}_\mathrm{pri}$, is implicitly constructed from these components:
\begin{equation}
\label{eq:w_primary_hat}
\widehat{\mathbf{W}}_\mathrm{pri} = \mathrm{diag}(\mathbf{h}) \mathbf{U}_\mathrm{sign} \mathrm{diag}(\boldell) \mathbf{V}_\mathrm{sign}^\top \mathrm{diag}(\mathbf{g})
\end{equation}
Instead of storing or optimizing $\widehat{\mathbf{W}}_\mathrm{pri}$ directly, the model learns latent full-precision factors $\mathbf{U}$ and $\mathbf{V}$. These are binarized during the forward pass using \cref{eq:uv_sign}, along with the FP16 scales $\mathbf{h}, \mathbf{g}, \boldell$. This factorized representation enables efficient computation during the forward pass.

\paragraph{Proposition 1}
Let $\mathbf{X} \in \mathbb{R}^{\mathrm{seq} \times d_\mathrm{in}}$ be the input matrix. The forward computation $\mathbf{Y} = \mathbf{X}{\widehat{\mathbf{W}}_\mathrm{pri}}^\top$ using the primary approximation (\cref{eq:w_primary_hat}) is efficiently computed as:
\begin{equation}
\label{eq:proposition1}
\mathbf{Y} = (( ( (\mathbf{X} \odot \mathbf{g}) \mathbf{V}_\mathrm{sign} ) \odot \boldell ) \mathbf{U}_\mathrm{sign}^\top ) \odot \mathbf{h}
\end{equation}
where $\odot$ denotes element-wise multiplication with broadcasting.

This decomposition (\cref{eq:proposition1}) replaces a large high-precision General Matrix Multiply (GEMM) operation with two smaller binary matrix multiplications and element-wise scaling operations. This offers computational advantages and reduces memory requirements. The effective bits per weight (BPW) is determined by the storage cost of these learnable parameters relative to the original matrix size (see \cref{appendix:average_bit} for details).

\subsection{Dual-SVID Initialization}
\label{sec:dual_svid}
Naively initializing the highly constrained \name structure can lead to unstable QAT. To circumvent this, we propose \textbf{Dual-SVID}, an SVD-based initialization method designed to provide a starting point where the initial effective weight $\widehat{\mathbf{W}}_{\mathrm{pri}, 0}$ closely approximates $\mathbf{W}$. Dual-SVID aims to preserve essential sign and magnitude information from the optimal low-rank SVD factors (obtained from $\mathbf{W} \approx \mathbf{U}' {\mathbf{V}'}^\top$, then truncated) and map this information to the learnable \name parameters ($\mathbf{U}_\mathrm{sign}, \mathbf{V}_\mathrm{sign}, \mathbf{h}, \mathbf{g}, \boldell$). The process involves three steps: (1) Obtaining $\mathbf{U}', \mathbf{V}'$ via a truncated SVD. (2) Setting initial binary factors based on signs (\cref{eq:sign_init_concise}). (3) Decomposing the magnitudes $|\mathbf{U}'| \in \mathbb{R}^{d_\mathrm{out} \times r}$ and $|\mathbf{V}'| \in \mathbb{R}^{d_\mathrm{in} \times r}$. To initialize the scales, a rank-1 SVD (or equivalent rank-1 approximation) is performed on each of these magnitude matrices separately. For $|\mathbf{U}'|$, its best rank-1 approximation can be written as $\mathbf{s}_{U,0} (\boldell_{u,0})^\top$, where $\mathbf{s}_{U,0} \in \mathbb{R}^{d_\mathrm{out}}$ becomes the initial row scale $\mathbf{h}_0$, and $\boldell_{u,0} \in \mathbb{R}^r$ is a component contributing to the latent scale. Similarly, for $|\mathbf{V}'|$, its rank-1 approximation $\mathbf{s}_{V,0} (\boldell_{v,0})^\top$ provides the initial column scale $\mathbf{g}_0 = \mathbf{s}_{V,0} \in \mathbb{R}^{d_\mathrm{in}}$ and another latent scale component $\boldell_{v,0} \in \mathbb{R}^r$. These steps effectively disentangle multi-dimensional magnitude variations and allow for the estimation of initial row ($\mathbf{h}_0$), column ($\mathbf{g}_0$), and the final latent scale ($\boldell_0 = \boldell_{u,0} \odot \boldell_{v,0}$ by element-wise product), as shown in \cref{eq:scale_init_rank1_concise,eq:latent_scale_init_concise_0}:
\begin{gather}
\label{eq:sign_init_concise}
\mathbf{U}_{\mathrm{sign}, 0} = \mathrm{sign}(\mathbf{U}'), \quad \mathbf{V}_{\mathrm{sign}, 0} = \mathrm{sign}(\mathbf{V}') \\
\label{eq:scale_init_rank1_concise}
|\mathbf{U}'| \approx \mathbf{h}_0 (\boldell_{u, 0})^\top, \quad |\mathbf{V}'| \approx \mathbf{g}_0 (\boldell_{v, 0})^\top \\
\label{eq:latent_scale_init_concise_0}
\boldell_0 = \boldell_{u, 0} \odot \boldell_{v, 0}
\end{gather}

The resulting initial effective weight $\widehat{\mathbf{W}}_{\mathrm{pri}, 0} = \mathrm{diag}(\mathbf{h}_0) \mathbf{U}_{\mathrm{sign}, 0} \mathrm{diag}(\boldell_0) \mathbf{V}_{\mathrm{sign}, 0}^\top \mathrm{diag}(\mathbf{g}_0)$ preserves key structural information. The learnable parameters are initialized as follows: the latent factors $\mathbf{U}$ and $\mathbf{V}$ are initialized with the factors $\mathbf{U}'$ and $\mathbf{V}'$ obtained from SVD, respectively, and the scales $\mathbf{h}, \mathbf{g}, \boldell$ are initialized to $\mathbf{h}_0, \mathbf{g}_0, \boldell_0$.

\subsection{Residual Compensation}
\label{sec:residual_compensation}
To further improve fidelity without increasing the primary path's rank, we introduce \textbf{Residual Compensation}. This technique is motivated by the theoretical insight that a two-stage error correction can be more effective than a single, larger approximation (\cref{appendix:proofs}). Crucially, this method does not increase the model's total parameter budget; instead, the fixed bit budget of a single, higher-rank approximation is strategically reallocated into two lower-rank paths: a primary and a residual path. The term `residual' primarily describes its role during initialization, where the auxiliary path is configured to model the error of the primary approximation. During QAT, both paths are optimized jointly to collectively represent the original weight. The advantage of this dual-path approach over a single path is empirically validated in our ablation studies (\cref{appendix:residual_compensation}).

\begin{figure}[t]
    \centering
    \vspace{10pt}
    \includegraphics[width=0.90\linewidth]{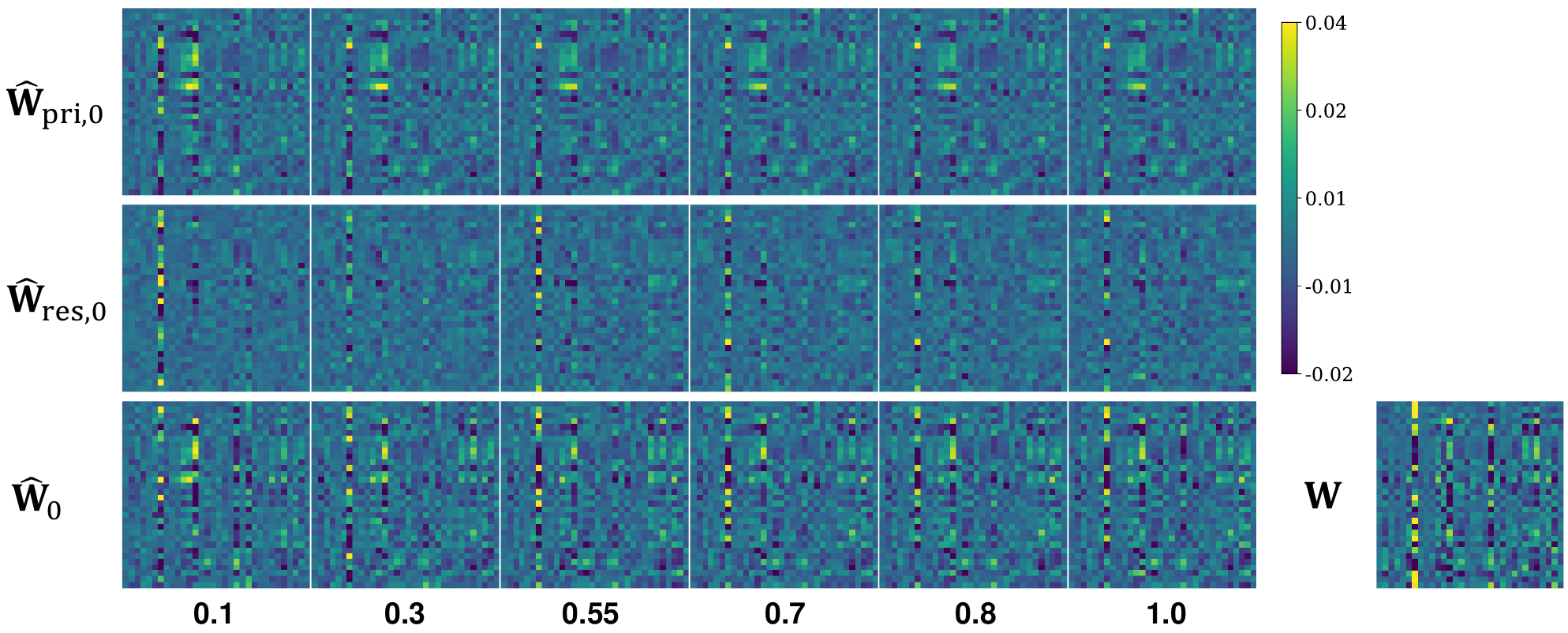}
    \caption{
         Visualization of Dual-SVID initialized weight components for a selected layer in Llama2-7B (Query Weight, Layer 0). Columns, from left to right, represent effective bits of 0.1, 0.3, 0.55, 0.7, 0.8, and 1.0 BPW. Rows display the primary approximation ($\widehat{\mathbf{W}}_{\mathrm{pri},0}$), the residual approximation ($\widehat{\mathbf{W}}_{\mathrm{res},0}$), and their sum ($\widehat{\mathbf{W}}_{0} = \widehat{\mathbf{W}}_{\mathrm{pri},0} + \widehat{\mathbf{W}}_{\mathrm{res},0}$). The rightmost image shows the corresponding crop of the original weight matrix ($\mathbf{W}$) for reference.
    }
    \label{fig:weight_heat}
    \vspace{-5pt}
\end{figure}

To implement this, Residual Compensation employs a parallel auxiliary path that learns an approximation $\widehat{\mathbf{W}}_\mathrm{res}$ of the residual error. This auxiliary path mirrors the structure of the \name path:
\begin{equation}
\label{eq:res_approx_structure}
\widehat{\mathbf{W}}_\mathrm{res} = \mathrm{diag}(\mathbf{h}_\mathrm{res}) \mathbf{U}_{\mathrm{res, sign}} \mathrm{diag}(\boldell_\mathrm{res}) \mathbf{V}_{\mathrm{res, sign}}^\top \mathrm{diag}(\mathbf{g}_\mathrm{res})
\end{equation}

The parameters of this auxiliary path ($\mathbf{U}_{\mathrm{res, sign}}, \mathbf{V}_{\mathrm{res, sign}}, \mathbf{h}_\mathrm{res}, \mathbf{g}_\mathrm{res}, \boldell_\mathrm{res}$) are also learnable during QAT. They are initialized using the Dual-SVID strategy applied to the initial residual error $\mathbf{W}_{\mathrm{res}, 0} = \mathbf{W} - \widehat{\mathbf{W}}_{\mathrm{pri}, 0}$. Employing a separate path allows the model to specifically learn and compensate for the potentially distinct characteristics of this residual error. The final effective weight is the sum of both approximations:
\begin{equation}
\label{eq:final_weight_approx_res}
\widehat{\mathbf{W}} = \widehat{\mathbf{W}}_\mathrm{pri} + \widehat{\mathbf{W}}_\mathrm{res}
\end{equation}

This targeted error correction is often advantageous for maintaining performance at extreme compression levels, as demonstrated by the ablation studies presented in \cref{appendix:residual_compensation}. It is important to note that all learnable parameters from both the primary and residual paths are included when calculating the final BPW (see \cref{appendix:average_bit} for calculation details).

\Cref{fig:weight_heat} visualizes the initial state of weight components ($\widehat{\mathbf{W}}_{\mathrm{pri},0}$, $\widehat{\mathbf{W}}_{\mathrm{res},0}$, and their sum $\widehat{\mathbf{W}}_{0}$) for a Llama2-7B query weight, derived via Dual-SVID across effective bits from 0.1 to 1.0, compared against the original weight $\mathbf{W}$ (far right). Even at a low 0.1 BPW, the primary approximation $\widehat{\mathbf{W}}_{\mathrm{pri},0}$ (top row, leftmost) captures the dominant low-rank structure of $\mathbf{W}$, albeit with considerable simplification. As effective bits increase, $\widehat{\mathbf{W}}_{\mathrm{pri},0}$ progressively refines details and more effectively suppresses disruptive approximation noise, allowing underlying weight patterns to emerge with greater clarity.
A key observation is that $\widehat{\mathbf{W}}_{0}$ (bottom row) at a modest 0.3 BPW---benefiting from the initialized residual component $\widehat{\mathbf{W}}_{\mathrm{res},0}$ (middle row)---often presents a more faithful initial representation of $\mathbf{W}$ than $\widehat{\mathbf{W}}_{\mathrm{pri},0}$ alone even at a higher 1.0 BPW.
This underscores that Residual Compensation, even at the initialization stage, is vital for capturing complexities missed by the primary low-rank approximation, thereby providing a more faithful starting point for QAT.

\section{Experiments}
\label{sec:experiments}

\subsection{Settings}
\label{sec:exp_settings}

\paragraph{Evaluation Setup}
We evaluated \name across diverse LLM families, including Llama~\citep{llama2023}, Llama2~\citep{llama2_2023}, Llama3~\citep{llama3_2024}, OPT~\citep{opt2022}, Phi-4~\citep{phi4_2024}, and QwQ~\citep{qwq32b2025}. These models span parameter scales from 1.3B to 32B. Perplexity (PPL) on the WikiText-2~\citep{wiki2_2016} validation dataset served as the primary performance metric. \Cref{appendix:ppl_c4_ptb} provides additional results on the C4~\citep{C42020} and PTB~\citep{ptb1994} datasets. Furthermore, we assessed zero-shot accuracy on common reasoning benchmarks: WinoGrande~\citep{winogrande2021}, OpenBookQA (OBQA)~\citep{OpenBookQA2018}, HellaSwag~\citep{hellaswag2019}, BoolQ~\citep{boolq2019}, ARC-Easy (ARC-e), ARC-Challenge (ARC-c)~\citep{arc2018}, and PIQA~\citep{piqa2020}.

\paragraph{Training Details}
We optimized the \name model parameters, initialized via the Dual-SVID method (\cref{sec:dual_svid}), using QAT with knowledge distillation (KD)~\citep{distilling2015,llmqat2023,ternarybert2020}. The original pre-trained full-precision model functioned as the teacher ($\mathcal{T}$) for the \name student model ($\mathcal{S}$). The QAT objective combines the standard output Kullback-Leibler (KL) divergence loss, $\mathcal{L}_\mathrm{out}$, and an intermediate layer mean squared error (MSE) loss, $\mathcal{L}_\mathrm{inter}$, to match hidden representations. We weighted these terms using an empirically determined coefficient $\lambda=10$:
\begin{equation}
\label{eq:KD_total_objective} 
\mathcal{L}_\mathrm{QAT} = \mathcal{L}_\mathrm{out} + \lambda \mathcal{L}_\mathrm{inter}.
\end{equation}
Adhering to the protocol in~\citep{binarymos2024}, the training data combined WikiText-2 with selected partitions from C4. The configuration included a sequence length of 2048 tokens, 5 epochs, the Adam optimizer ($\beta_1 = 0.9, \beta_2 = 0.999$)~\citep{adam2014}, and a cosine learning rate decay with 2\% warm-up (see \cref{appendix:hyper_parameters} for details). For models employing Grouped-Query Attention (GQA)~\citep{gqa2023}, such as Llama3, Phi-4, and QwQ, we specifically adjusted the latent ranks $r$ for key (K) and value (V) projection layers (see \cref{appendix:gqa}) to maintain performance under extreme quantization. We employed \textit{SmoothSign} (forward: $\mathrm{sign}(x)$; backward gradient: $\tanh(kx)$, $k=100$) for backpropagation, as it demonstrated superior performance compared to the Straight-Through Estimator (STE)~\citep{ste2013} (see \cref{appendix:ste}). Although proxy gradients are established in the literature, this work validates the specific application of the $\tanh(100x)$ derivative as a stable gradient for QAT in the ultra-low-bit regime. Our ablation study confirms the effectiveness of this approach (\cref{appendix:ste}).

\paragraph{Baselines}
We benchmarked \name across effective bit rates ranging from approximately 1.0 down to 0.1 bits per weight (BPW). In the challenging sub-1-bit regime, we compared our method against STBLLM~\citep{stbllm2024}, a post-training quantization (PTQ) method that employs N:M sparsity. For settings approximating 1.0 BPW, we compared against the 1-bit PTQ method BiLLM~\citep{billm2024} and quantization-aware training (QAT) approaches such as OneBit~\citep{onebit2024} and BinaryMoS~\citep{binarymos2024}. We included results from standard low-bit methods, such as GPTQ~\citep{gptq2023} and OmniQuant~\citep{omniquant2023}, to provide broader context.

\subsection{Main Results}
\paragraph{Superior Perplexity in Sub-1-bit Regime}
\Cref{tab:main_results} presents the perplexity (PPL) results of \name compared to baseline methods. The proposed method demonstrates robust performance, particularly within the sub-1-bit regime. Relative to STBLLM~\citep{stbllm2024}, \name achieves markedly superior PPL scores at $0.8$, $0.7$, and $0.55$ BPW. For instance, on Llama2-7B, \name at $0.55$ BPW achieves a PPL of $10.47$, which represents a substantial improvement over the $30.67$ score reported for STBLLM. We observe comparable gains across other models and bit-widths. For Llama2-13B at $0.8$ BPW, \name records a PPL of $8.52$, whereas STBLLM records $11.90$.

\begin{table*}[t]
    \renewcommand{\arraystretch}{1.0}
    \small
    \centering
    \setlength{\tabcolsep}{1.50mm}
    \caption{
            Perplexity (PPL) comparison on WikiText-2 across various LLMs and quantization methods. Lower PPL indicates better performance. BPW denotes effective bits per weight. Methods marked with $^\dagger$ use quantization-aware training (QAT). STBLLM~\citep{stbllm2024} utilizes N:M sparsity (ratio in parentheses). \name demonstrates strong performance, particularly in the sub-1-bit regime.
    }
    \label{tab:main_results}
    \resizebox{0.90\textwidth}{!}{
        \begin{tabular}{lccrrrrrrrr}
        \toprule
        \multicolumn{3}{c}{\textbf{Settings}} & \multicolumn{1}{c}{\textbf{ OPT }} & \multicolumn{2}{c}{\textbf{ Llama }}  & \multicolumn{2}{c}{\textbf{ Llama2 }} & \multicolumn{1}{c}{\textbf{ Llama3}} & \multicolumn{1}{c}{\textbf{ Phi-4 }}   & \multicolumn{1}{c}{\textbf{ QwQ}} \\ 
        \midrule
        Method & Block Size & BPW & 1.3B    & 7B   & 13B    & 7B   & 13B   & 8B    & 14.7B & 32B \\ 
        \midrule
        FullPrecision & -     & 16    & 14.62  & 5.67  & 5.09    & 5.47  & 4.88  & 6.10  & 6.67 & 6.34 \\ 
        OmniQuant     &     - & 2     & 28.82 & 9.75 & 7.83 & 11.20 & 8.25 & 349.27 & 12.09  & 10.19 \\ 
        GPTQ          & 128     & 2 & 119.98     & 39.96 & 15.08       & 52.22         & 23.63         & 1.5e3         & 24.96 & 67.16 \\
        BiLLM & 128   & 1.1  & 74.07 & 41.95 & 13.95 & 29.00  & 23.71 & 54.29 & 16.95 & 15.4 \\ 
        OneBit$^\dagger$ & -   & 1   & 20.27 & 8.21 & 7.37 & 8.36  & 7.41 & 13.09 & 9.92 & 9.86 \\
        BinaryMoS$^\dagger$ & -    & 1  & 18.09 & 7.84 & 7.05 & 7.74  & 6.95 & 10.83 & 9.51 & 8.99 \\ 
        \name$^\dagger$ & -    & 1     & 20.33 & 9.03 & 8.17 & 9.08  & 8.18 & 16.30 & 11.28 & 12.08 \\ 
        \midrule
        STBLLM & 128    & 0.80 (6:8)     & 52.13 & 15.19 & 9.43 & 13.81    & 11.90 & 30.90 & 12.12 & 12.19 \\ 
        STBLLM & 128    & 0.70 (5:8)     & 73.42 & 19.52 & 11.47 & 19.17    & 14.00 & 59.83 & 14.57 & 13.78 \\ 
        STBLLM & 128    & 0.55 (4:8)     & 123.03 & 38.73 & 16.88 & 30.67    & 27.05 & 241.95 & 21.99 & 18.32 \\ 
        STBLLM & 128    & 0.30 (2:8)     & 2.3e3 & 1.6e3 & 592.06 & 1.8e3    & 893.82 & 1.7e5 & 761.05 & 512.01\\ 
        \midrule
        \name$^\dagger$ & -    & 0.80     & 21.32 & 9.44 & 8.50 & 9.53    & 8.52 & 17.28 & 11.70 & 12.46\\ 
        \name$^\dagger$ & -    & 0.70     & 21.99 & 9.66 & 8.71 & 9.85    & 8.76 & 18.01 & 12.05 & 13.22 \\ 
        \name$^\dagger$ & -    & 0.55     & 23.35 & 10.09 & 9.16 & 10.47    & 9.24 & 18.47 & 12.80 & 13.57\\ 
        \name$^\dagger$ & -    & 0.30     & 28.30 & 11.50 & 10.33 & 12.00    & 10.48 & 20.34 & 14.71 & 16.48\\
        \name$^\dagger$ & -    & 0.10     & 53.76 & 15.58 & 13.71 & 15.92    & 15.09 & 26.11 & 19.73 & 35.26 \\ 
        \bottomrule
        \end{tabular}
    }
\end{table*}

\begin{table*}[t]
  \centering
    \caption{
    Zero-shot accuracy (\%) comparison on common sense reasoning benchmarks. Compares Full Precision (FP16) models against STBLLM~\citep{stbllm2024} and \name at 0.55 and 0.3 BPW.}
    \label{tab:zero_shot_main_detail}
    \resizebox{0.90\textwidth}{!}{
      \setlength{\tabcolsep}{5.5pt}
        \begin{tabular}{llcccccccc}
        \toprule
        \textbf{Models} & \textbf{Method} & \textbf{WinoGrande} & \textbf{OBQA}  & \textbf{HellaSwag} & \textbf{BoolQ} & \textbf{ARC-e}  & \textbf{ARC-c}  & \textbf{PIQA}   & \textbf{Average} \\
        \midrule
        \multirow{5}[2]{*}{Llama2-7B} & FullPrecision & 67.16  & 40.80  & 72.95  & 71.37  & 69.44  & 40.95  & 78.12  & 62.97  \\
              & STBLLM (0.55) & 52.80 & 33.00 & 36.94 & 62.17 & 37.33 & 25.34 & 62.46 & 44.29  \\
              & STBLLM (0.3) & 50.43 & 31.80 & 26.20 & 37.83 & 25.88 & 25.09 & 53.26 & 35.78  \\
              & \name (0.55) & 51.62 & 34.00 & 44.57 & 61.38 & 44.15 & 26.96 & 68.12 & \textbf{47.26}  \\
              & \name (0.3) & 51.30 & 34.80 & 37.61 & 61.80 & 39.98 & 25.09 & 65.83 & \textbf{45.20}  \\
        \midrule
        \multirow{5}[2]{*}{Llama2-13B} & FullPrecision & 69.45  & 41.80  & 76.58  & 69.29  & 73.19  & 44.53  & 78.61  & 64.78  \\
              & STBLLM (0.55) & 55.25 & 31.20 & 35.10 & 62.23 & 42.51 & 27.22 & 61.75 & 45.04  \\
              & STBLLM (0.3) & 51.54 & 28.80 & 26.01 & 55.63 & 27.53 & 24.74 & 53.32 & 38.22  \\
              & \name (0.55) & 53.04 & 35.60 & 51.06 & 50.58 & 46.09 & 29.10 & 70.78 & \textbf{48.03}  \\
              & \name (0.3) & 51.30 & 34.00 & 43.81 & 55.96 & 42.59 & 25.17 & 68.77 & \textbf{45.94}  \\
        \midrule
        \multirow{5}[2]{*}{Llama3-8B} & FullPrecision & 72.92  & 45.00  & 79.18  & 81.25  & 80.21  & 52.98  & 79.54  & 70.15  \\
              & STBLLM (0.55) & 52.17 & 25.80 & 30.61 & 57.16 & 30.35 & 23.72 & 57.56 & 39.62 \\
              & STBLLM (0.3) & 49.57 & 26.60 & 26.36 & 51.62 & 26.05 & 24.40 & 51.74 & 36.62 \\
              & \name (0.55) & 50.12 & 30.20 & 36.78 & 57.55 & 46.80 & 22.95 & 66.27 & \textbf{44.38}  \\
              & \name (0.3) & 51.93  & 28.20  & 33.91  & 57.98  & 43.48  & 24.32  & 64.64  & \textbf{43.49}  \\
        \bottomrule
        \end{tabular}%
    }
\end{table*}

\begin{figure}[t]
    \centering
    \begin{subfigure}{0.48\linewidth}
        \centering
        \includegraphics[width=\linewidth]{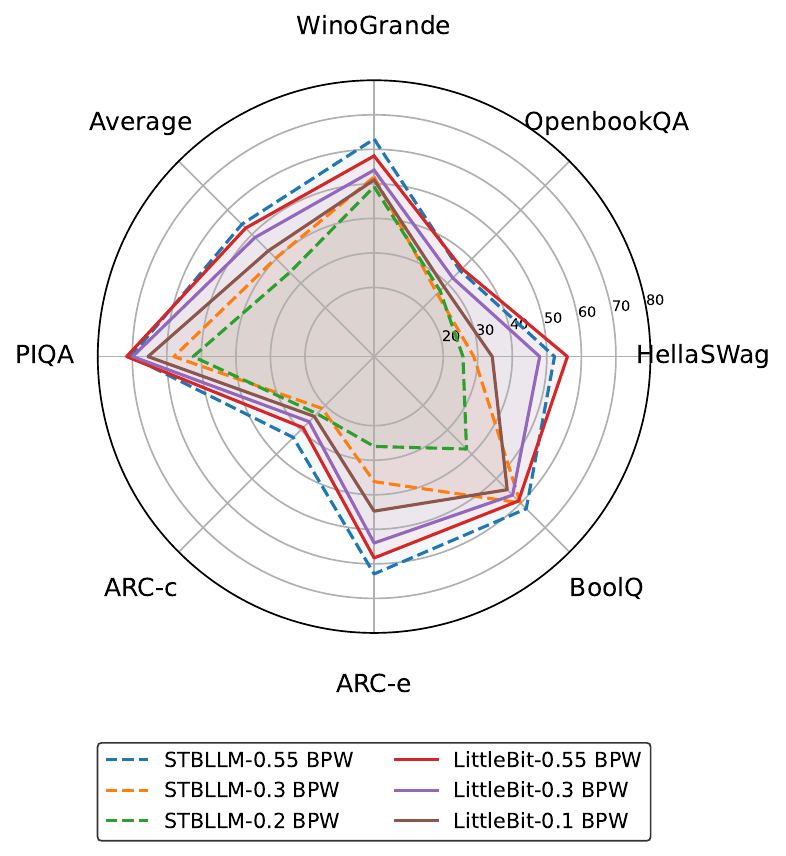}
        \caption{Phi-4 14B}
        \label{fig:zeroshot_phi4}
    \end{subfigure}
    \hfill
    \begin{subfigure}{0.48\linewidth}
        \centering
        \includegraphics[width=\linewidth]{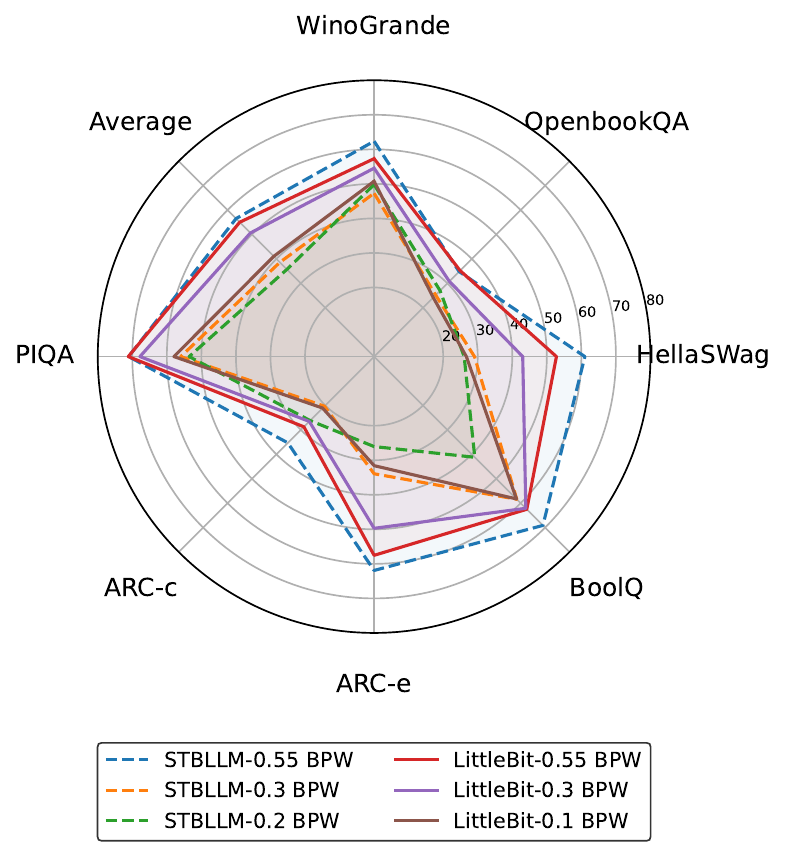}
        \caption{QwQ-32B}
        \label{fig:zeroshot_qwq}
    \end{subfigure}
    \caption{Zero-shot accuracy (\%) on 7 common sense reasoning tasks. The comparison highlights the robustness of \name (solid lines) against STBLLM (dashed lines) in the sub-0.5 BPW regime. While STBLLM degrades sharply, \name preserves viable reasoning capabilities for both (a) Phi-4 and (b) QwQ models.}
    \label{fig:zeroshot_results}
\end{figure}

\paragraph{Extreme Low-BPW Stability}
A primary advantage of \name lies in its exceptional stability and performance in the extreme sub-0.5 BPW range, specifically at $0.3$ BPW and $0.1$ BPW. In this regime, methods such as STBLLM exhibit severe performance degradation. For example, Llama2-7B yields a PPL of $1.8 \times 10^3$ at $0.3$ BPW. Conversely, \name maintains strong PPL scores, such as $12.00$ for Llama2-7B at $0.3$ BPW and $15.92$ at an unprecedented rate of $0.1$ BPW. When configured for effective bits approximating $1.0$ BPW by adjusting the latent rank $r$, \name yields perplexity scores comparable to specialized 1-bit QAT methods. For example, on Llama2-7B, \name achieves $9.08$ compared to $8.36$ for OneBit~\citep{onebit2024}. It also remains competitive with PTQ methods such as BiLLM~\citep{billm2024}. Although BinaryMoS~\citep{binarymos2024} may exhibit slightly lower PPL in certain 1-bit scenarios (\eg Llama2-7B: BinaryMoS $7.74$), this performance difference is attributable to its use of more complex mechanisms, such as dynamic scaling, which represent a distinct set of architectural and computational trade-offs. Our analysis of the trade-off involved in adjusting the latent rank to tune effective bits from $1.0$ down to $0.1$ BPW indicates that performance degradation is minimal down to approximately $0.55$ BPW. However, a quantization cliff appears between $0.3$ and $0.1$ BPW. This observation positions the $0.3$--$0.55$ BPW range as an optimal sweet spot for balancing compression and accuracy.

\paragraph{Model Generalization \& Zero-Shot}
We consistently observe the advantages of \name across a diverse set of model architectures and sizes. This includes the Llama3 family, which is typically considered challenging for quantization, as well as other contemporary models such as Phi-4. For example, Llama3-8B quantized by \name to $0.55$ BPW yields a PPL of $18.47$. This result substantially outperforms the STBLLM score of $241.95$ at $0.55$ BPW with 4:8 sparsity. The zero-shot evaluation results presented in \cref{tab:zero_shot_main_detail} corroborate these perplexity-based findings. For instance, Llama2-7B compressed by \name to $0.55$ BPW achieves a mean accuracy of $47.26\%$, and the model maintains $45.20\%$ mean accuracy even at an extreme $0.3$ BPW. These results compare favorably with those of STBLLM at similar or even higher effective bits. STBLLM on Llama2-7B at $0.55$ BPW scores $44.29\%$. This outcome suggests superior preservation of the intrinsic reasoning capabilities of the models even under extreme compression. It highlights the efficacy of \name in creating highly compact yet powerful LLMs.

\subsection{Reasoning Performance of Extremely Compressed LLMs}
The capability of \name to achieve sub-1-bit compression while maintaining robust model fidelity prompts an evaluation of its reasoning abilities, particularly for large-scale models under extreme compression. We investigated the Phi-4 14B model and extended our analysis to the QwQ-32B model, comparing their performance against STBLLM across 7 zero-shot reasoning tasks.

When compressed with \name to $0.55$ BPW, Phi-4 achieved an average accuracy of approximately $52.3\%$. Although this performance is marginally lower than STBLLM at a similar compression level ($54.2\%$ at $0.55$ BPW), \name demonstrates substantially more graceful degradation. At the more aggressive $0.3$ BPW setting, the \name-compressed Phi-4 maintained an average accuracy of $48.7\%$, which markedly outperforms STBLLM at $0.3$ BPW ($40.3\%$). Even at an extreme $0.1$ BPW, Phi-4 retained a notable accuracy of $43.6\%$. We observed a consistent trend with the QwQ-32B model, as illustrated in \Cref{fig:zeroshot_results}. While STBLLM exhibited severe performance collapse in the sub-0.5 BPW regime, \name maintained a stable trajectory down to 0.1 BPW. These results highlight the ability of \name to preserve robust reasoning capabilities across varying model scales as compression becomes increasingly extreme.

\begin{table*}[t]
    \renewcommand{\arraystretch}{1.0}
    \small
    \centering
    \setlength{\tabcolsep}{1.50mm}
    \caption{Memory footprint comparison (GB) for Llama2~\citep{llama2_2023} models under different quantization methods. \name is evaluated at various BPWs, achieved by adjusting the latent rank $r$. Compression factors relative to FP16 are shown in parentheses.The gap between the effective BPW and the compression factor arises because quantization is applied only to the Transformer blocks, while components like the embedding layer and the lm\_head remain in FP16.}
    \label{tab:memory_footprint}
    \resizebox{0.95\textwidth}{!}{
    \begin{tabular}{lccccccc}
        \toprule
        Model       & FP16   & BiLLM~\citep{billm2024} & OneBit~\citep{onebit2024} & \multicolumn{4}{c}{\textbf{ \name (Ours)}} \\
        \midrule
        BPW & 16 & 1.1 & 1 & 0.8 & 0.55 & 0.3 & 0.1  \\
        \midrule
        Llama2-7B   & 13.49\,GB & 1.60\,GB (8.43$\times$) & 1.36\,GB (9.92$\times$) & 1.19\,GB (11.34$\times$) & 0.98\,GB (13.77$\times$) & 0.79\,GB (17.08$\times$) & 0.63\,GB (21.41$\times$) \\
        Llama2-13B  & 26.06\,GB & 2.80\,GB (9.31$\times$) & 2.28\,GB (11.43$\times$) & 1.95\,GB (13.36$\times$) & 1.51\,GB (17.26$\times$) & 1.15\,GB (22.66$\times$) & 0.84\,GB (31.02$\times$) \\
        Llama2-70B  & 138.04\,GB & 15.40\,GB (8.96$\times$) & 9.72\,GB (14.20$\times$) & 7.97\,GB (17.31$\times$) & 5.83\,GB (23.68$\times$) & 3.70\,GB (37.31$\times$) & 1.98\,GB (69.72$\times$) \\
        \bottomrule
    \end{tabular}}
    \vspace{-10pt}
\end{table*}
\begin{figure}[t]
  \centering
  \begin{minipage}[ht]{0.48\linewidth}
    \centering
    \includegraphics[width=0.98\linewidth]{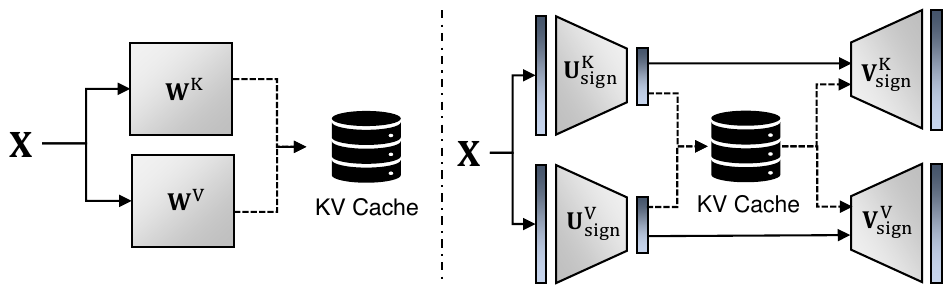}
    \captionof{figure}{Conceptual view of KV Cache storage: the standard method (left) stores the full hidden dimension ($d_\mathrm{model}$), whereas \name (right) caches a reduced latent dimension ($r$)}
    \label{fig:kv_cache_viz} 
  \end{minipage}
  \hfill
  \begin{minipage}[ht]{0.48\linewidth}
    \centering 
    \captionof{table}{Estimated KV cache memory reduction for Llama2-7B using \name, with a reduction factor of approximately $d_\mathrm{model}/r$.}
    \resizebox{0.85\textwidth}{!}{
      \begin{tabular}{ccc}
        \toprule
        BPW  & KV Latent Rank & KV Cache \\ 
          &  ($r$) & Reduction Factor \\
        \midrule
        0.80  & 1\small{,}624 & $\approx$2.5$\times$ \\
        0.55  & 1\small{,}112 & $\approx$3.7$\times$ \\
        0.30  & 600 & $\approx$6.8$\times$ \\
        0.10  & 192 & $\approx$21.3$\times$ \\
        \bottomrule 
      \end{tabular}
    }
    \label{tab:kv_cache_reduction}
  \end{minipage}
  \vspace{-5pt}
\end{figure}

\section{Analysis}
\label{sec:analysis}
\vspace{-10pt}
\paragraph{Memory Footprint Reduction}
\name is designed for resource-constrained environments, such as on-device deployment, and achieves substantial reductions in model memory footprint. As summarized in \cref{tab:memory_footprint}, quantizing a Llama2-7B model (originally $13.49$ GB in FP16) to an effective bit rate of $0.3$ BPW using \name reduces its required storage to $0.79$ GB. This corresponds to a compression factor exceeding $17\times$. At the extreme setting of $0.1$ BPW, the footprint is further reduced to $0.63$ GB (over $21\times$ compression). We observe comparable reductions for larger models, such as Llama2-70B, where the memory footprint decreases from $138.04$ GB to under $2$ GB at $0.1$ BPW, achieving nearly $70\times$ compression. These substantial memory reductions significantly expand the range of devices capable of hosting large-scale language models, thereby enabling complex language tasks on hardware with limited VRAM or storage capacity.

\paragraph{KV Cache Compression}
The proposed factorization (\cref{eq:w_primary_hat}) inherently compresses the KV cache when applied to key and value projection matrices. Because the forward computation (\cref{eq:proposition1}) operates on latent states, intermediate rank $r$ vectors are cached rather than full ${d}_\mathrm{model}$-dimensional vectors, as illustrated in \cref{fig:kv_cache_viz}.  Consequently, KV cache memory requirements are reduced by a factor of approximately ${d}_\mathrm{model}/{r}$. For example, a reduction of up to 21.3$\times$ is observed for Llama2-7B at 0.1 BPW (\cref{tab:kv_cache_reduction}). While this effect aligns with explicit compression methods such as MLA~\citep{deepseekv2_2024} and ASVD~\citep{asvd2023}, the approach described herein achieves this via unified factorization within attention layers, thereby reducing both weight and activation memory simultaneously.

\begin{figure}[t]
    \centering 
    \includegraphics[width=0.7\linewidth]{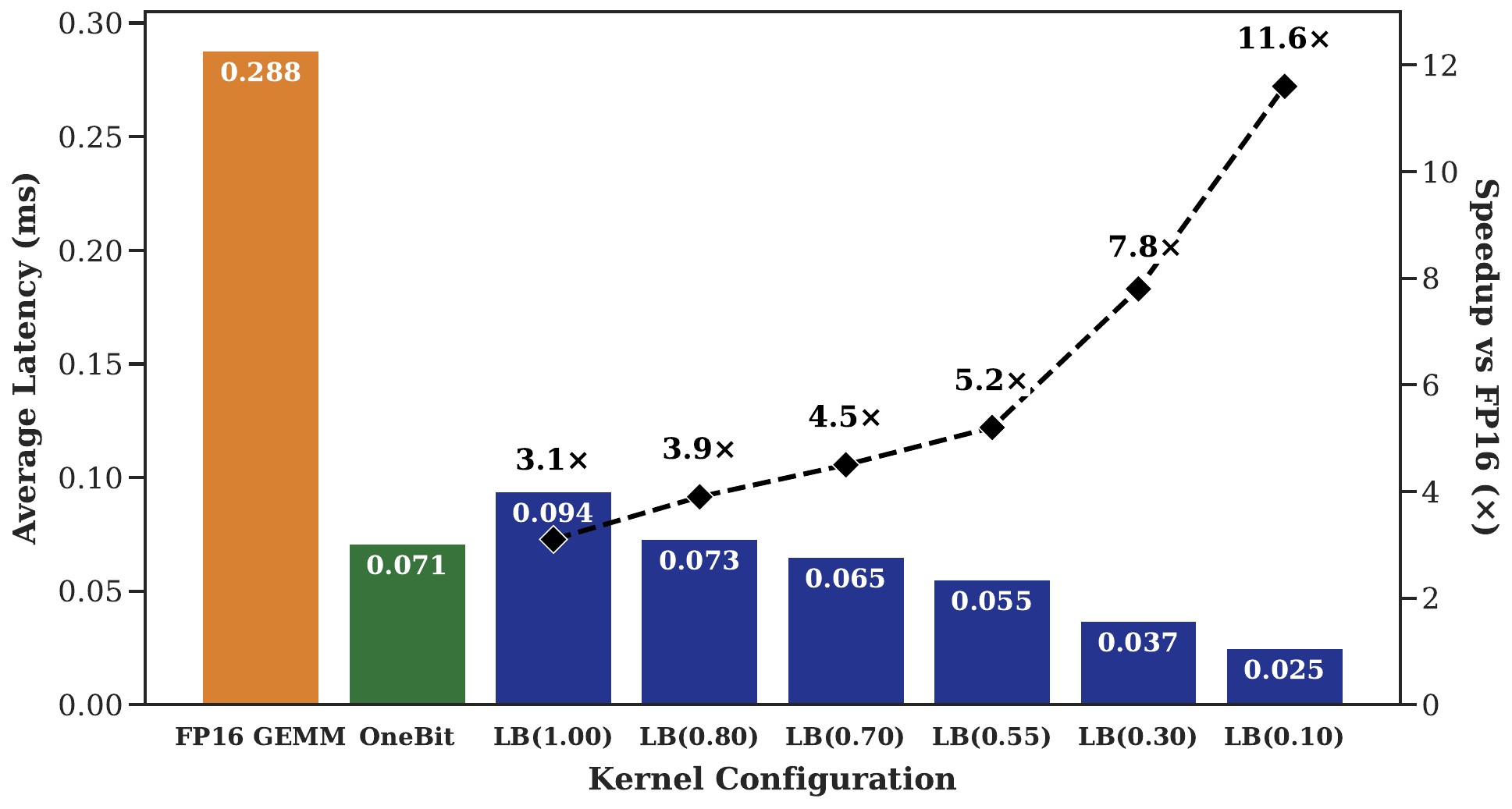}
    \captionof{figure}{Kernel-level latency (bars, left axis) and speedup relative to FP16 (dashed line, right axis) on an NVIDIA A100 GPU for a Llama2-70B MLP layer ($8,192\times 28,672$). The benchmark compares the FP16 GEMM baseline with OneBit and the proposed \name (LB) kernel across various BPW configurations ranging from 1.0 to 0.1.}
    \label{fig:latency_results}
    \vspace{-5pt}
\end{figure}

\paragraph{Latency Considerations}
\label{sec:latency}
In addition to memory optimization, inference latency presents a critical performance metric.  The factorized architecture (\cref{eq:proposition1}) facilitates computational acceleration through low-rank binary operations, distinct from methods solely prioritizing memory or requiring specific hardware. To evaluate this capability, a custom 1-bit GEMV CUDA kernel was developed to accelerate the primary computational path. As illustrated in \cref{fig:latency_results} for a Llama2-70B MLP layer, the kernel demonstrates speedups that inversely correlate with bit-width, peaking at \textbf{11.6$\times$} over an optimized FP16 baseline at 0.1 BPW.

These findings suggest the proposed method enables efficient deployment with reduced latency. The discrepancy between theoretical computational gains and practical latency is likely attributable to memory access dominance during small-batch inference. The factorized approach involves multiple memory-bound operations, and the custom kernel has not yet reached the optimization level of industry-standard libraries. \Cref{appendix:latency} provides a comprehensive analysis covering theoretical FLOPs, practical latency, and end-to-end decoding throughput.

\section{Conclusion}
\label{sec:conclusion}
This study proposed \name, a method designed to advance LLM compression into the sub-0.5 BPW regime while maintaining viable performance at 0.1 BPW. Performance improvements were realized through SVD-inspired latent matrix factorization, factor binarization, and a multi-scale compensation mechanism encompassing row, column, and latent dimensions. Furthermore, Dual-SVID initialization was introduced to stabilize quantization-aware training (QAT), alongside Residual Compensation for error mitigation. Empirical results demonstrate that the proposed method markedly outperforms prior sub-1-bit techniques and preserves fidelity across LLMs scaling up to 32B parameters. Consequently, \name suggests a favorable size–performance trade-off, thereby facilitating LLM deployment in resource-constrained environments. Future research may address practical deployment through hardware co-design for edge devices, such as neural processing units (NPUs). Additional improvements in model fidelity could be examined by integrating advanced techniques such as non-uniform bit allocation, applied either across model layers or within single layers via dynamic bit budget distribution between primary and residual paths.

\section*{Acknowledgements}
We are grateful to Dr. Heonjae Ha, Dr. SangJeong Lee, Minseop Choi, Changdong Kim, and Hyochan Chong for their insightful discussions and generous support in this work.

\def\bibfont{\small}

\medskip
\bibliographystyle{unsrtnat}
\bibliography{neurips_2025}

\newpage
\appendix

\crefalias{section}{appendix}
\crefalias{subsection}{appendix}
\crefname{appendix}{Appendix}{Appendices}

\section{Mathematical Analysis and Proofs}
\label{appendix:proofs}
This section provides mathematical details for the \name computation and explores theoretical aspects related to low-rank quantization errors, offering motivation for certain design choices.

\begin{proposition}[\name Forward Pass Computation (Proof Detail)]
\label{prop:forward_pass}
As stated in \cref{sec:little_bit_arch} (Proposition 1), for input $\mathbf{X} \in \mathbb{R}^{\mathrm{seq} \times d_{\mathrm{in}}}$, the primary quantized weight matrix is $\widehat{\mathbf{W}}_{\mathrm{pri}} = \mathrm{diag}(\mathbf{h}) \mathbf{U}_{\mathrm{sign}} \mathrm{diag}(\boldell) \mathbf{V}_{\mathrm{sign}}^\top \mathrm{diag}(\mathbf{g})$. The forward pass $\mathbf{Y} = \mathbf{X} \widehat{\mathbf{W}}_{\mathrm{pri}}^\top$ can be computed as shown in \cref{eq:forward_decomp_appendix_revised}.
\begin{equation}
\label{eq:forward_decomp_appendix_revised}
\mathbf{Y} = (( ( (\mathbf{X} \odot \mathbf{g}) \mathbf{V}_\mathrm{sign} ) \odot \boldell ) \mathbf{U}_\mathrm{sign}^\top ) \odot \mathbf{h}.
\end{equation}
Terms are defined in \cref{sec:little_bit_arch}.
\end{proposition}
\begin{proof}
Starting with $\mathbf{Y} = \mathbf{X} \widehat{\mathbf{W}}_{\mathrm{pri}}^\top$:
\begin{align}
\mathbf{Y}
&= \mathbf{X} \left( \mathrm{diag}(\mathbf{h}) \mathbf{U}_{\mathrm{sign}} \mathrm{diag}(\boldell) \mathbf{V}_{\mathrm{sign}}^\top \mathrm{diag}(\mathbf{g}) \right)^\top \nonumber \\
&= \mathbf{X} \, \mathrm{diag}(\mathbf{g}) \mathbf{V}_{\mathrm{sign}} \mathrm{diag}(\boldell) \mathbf{U}_{\mathrm{sign}}^\top \mathrm{diag}(\mathbf{h})  \nonumber \\
&= (( ( (\mathbf{X} \odot \mathbf{g}) \mathbf{V}_\mathrm{sign} ) \odot \boldell ) \mathbf{U}_\mathrm{sign}^\top ) \odot \mathbf{h}. \label{eq:proof_final_step_condensed}
\end{align}
This yields \cref{eq:forward_decomp_appendix_revised}, where $\odot$ implies element-wise multiplication with broadcasting.
\end{proof}

\begin{claim}[Quantization Error vs. Factor Rank]
\label{claim:monotonic_error}
Consider a simplified model where a rank-$r$ matrix $\mathbf{W}^{(r)} = \mathbf{U}^{(r)} (\mathbf{V}^{(r)})^\top$ is approximated. Let $\mathbf{U}_{\mathrm{sign}}^{(r)} = \mathrm{sign}(\mathbf{U}^{(r)})$, $\mathbf{V}_{\mathrm{sign}}^{(r)} = \mathrm{sign}(\mathbf{V}^{(r)})$. If magnitudes are crudely approximated using fixed rank-1 SVDs of $|\mathbf{U}^{(r)}|$ and $|\mathbf{V}^{(r)}|$ to form scales $\mathbf{s}_U^{(r)}, \mathbf{s}_V^{(r)}$:
\begin{equation}
\label{eq:W_hat_rank1_appendix_clarified_condensed}
\widehat{\mathbf{W}}_r :=
\left( \mathbf{U}_{\mathrm{sign}}^{(r)} \odot (\mathbf{s}_U^{(r)} \mathbf{1}^\top_r) \right)
\left( \mathbf{V}_{\mathrm{sign}}^{(r)} \odot (\mathbf{s}_V^{(r)} \mathbf{1}^\top_r) \right)^\top.
\end{equation}
The error $\mathcal{E}(r) := \left\| \mathbf{W}^{(r)} - \widehat{\mathbf{W}}_r \right\|_F$ might be non-decreasing with $r$.
\end{claim}
\begin{proof}[Proof Sketch / Heuristic Argument]
As rank $r$ (and thus complexity of $\mathbf{W}^{(r)}$, $\mathbf{U}^{(r)}$, $\mathbf{V}^{(r)}$) increases, the fixed-structure rank-1 magnitude approximation (via $\mathbf{s}_U^{(r)}, \mathbf{s}_V^{(r)}$) may become a relatively poorer fit for the increasingly complex actual magnitudes of $\mathbf{U}^{(r)}, \mathbf{V}^{(r)}$. This degradation in scaling accuracy could lead to $\mathcal{E}(r)$ not decreasing, or even increasing, motivating more sophisticated scaling as in \name.
\end{proof}

\begin{claim}[Quantization Bias vs. SVD Component Structure]
\label{claim:residual_bias}
Let $\mathbf{W}$'s rank-$r$ SVD approximation be $\widetilde{\mathbf{W}}_r = \widetilde{\mathbf{W}}_{r_1} + \widetilde{\mathbf{R}}_{r_2}$ (primary $\widetilde{\mathbf{W}}_{r_1}$, residual $\widetilde{\mathbf{R}}_{r_2}$). Consider a quantization operator $\mathcal{Q}$. The relative quantization error $\|\mathcal{Q}(\mathbf{A}) - \mathbf{A}\|_F / \|\mathbf{A}\|_F$ might be larger for matrices $\mathbf{A}$ with "flatter" singular value spectra or less energy concentration (\eg $\widetilde{\mathbf{R}}_{r_2}$ vs. $\widetilde{\mathbf{W}}_{r_1}$). This suggests that $\mathcal{Q}$ applied to $\widetilde{\mathbf{R}}_{r_2}$ might be less effective (\ie higher relative quantization error) than when applied to $\widetilde{\mathbf{W}}_{r_1}$.
\end{claim}
\begin{proof}[Proof Sketch]
$\widetilde{\mathbf{W}}_{r_1}$ has dominant singular values, while $\widetilde{\mathbf{R}}_{r_2}$ has smaller ones, often resulting in lower energy concentration for $\widetilde{\mathbf{R}}_{r_2}$. If $\mathcal{Q}$'s relative error is sensitive to this (\eg higher for lower concentration), quantizing $\widetilde{\mathbf{W}}_{r_1}$ and $\widetilde{\mathbf{R}}_{r_2}$ separately may be better than quantizing $\widetilde{\mathbf{W}}_r$ jointly, as it allows adapting $\mathcal{Q}$ to their differing structures.
\end{proof}

\begin{proposition}[Potential Advantage of Two-Stage Quantization of SVD Components]
\label{prop:two_stage}
Using the setup of \cref{claim:residual_bias}, let $\widehat{\mathbf{W}}_r^{\mathrm{single}} = \mathcal{Q}(\widetilde{\mathbf{W}}_r)$ and $\widehat{\mathbf{W}}^{\mathrm{two-stage}} = \mathcal{Q}(\widetilde{\mathbf{W}}_{r_1}) + \mathcal{Q}(\widetilde{\mathbf{R}}_{r_2})$. The two-stage error $\mathcal{E}_{\mathrm{two-stage}} = \left\| \mathbf{W} - \widehat{\mathbf{W}}^{\mathrm{two-stage}} \right\|_F$ can be less than the single-stage error $\mathcal{E}_{\mathrm{single-stage}} = \left\| \mathbf{W} - \widehat{\mathbf{W}}_r^{\mathrm{single}} \right\|_F$. This occurs if the quantization error of the sum is greater than the sum of (vectorial) quantization errors of parts:
\begin{equation}
\label{eq:quant_error_cond_condensed}
\| (\mathcal{Q}(\widetilde{\mathbf{W}}_{r_1}) - \widetilde{\mathbf{W}}_{r_1}) + (\mathcal{Q}(\widetilde{\mathbf{R}}_{r_2}) - \widetilde{\mathbf{R}}_{r_2}) \|_F < \| \mathcal{Q}(\widetilde{\mathbf{W}}_r) - \widetilde{\mathbf{W}}_r \|_F.
\end{equation}
\end{proposition}
\begin{proof}[Proof Outline and Discussion]
Let $\mathbf{E}_{\mathrm{trunc}} = \mathbf{W} - \widetilde{\mathbf{W}}_r$. Errors are $\Delta_1 = \mathcal{Q}(\widetilde{\mathbf{W}}_{r_1}) - \widetilde{\mathbf{W}}_{r_1}$, $\Delta_2 = \mathcal{Q}(\widetilde{\mathbf{R}}_{r_2}) - \widetilde{\mathbf{R}}_{r_2}$, $\Delta_{\mathrm{sum}} = \mathcal{Q}(\widetilde{\mathbf{W}}_r) - \widetilde{\mathbf{W}}_r$.
Then $\mathcal{E}_{\mathrm{two-stage}} = \|\mathbf{E}_{\mathrm{trunc}} - (\Delta_1 + \Delta_2)\|_F$ and $\mathcal{E}_{\mathrm{single-stage}} = \|\mathbf{E}_{\mathrm{trunc}} - \Delta_{\mathrm{sum}}\|_F$.
The condition \cref{eq:quant_error_cond_condensed} makes $\mathcal{E}_{\mathrm{two-stage}} < \mathcal{E}_{\mathrm{single-stage}}$ likely. Non-linear $\mathcal{Q}$ (\eg with adaptive scaling) can satisfy this by better handling the distinct characteristics of $\widetilde{\mathbf{W}}_{r_1}$ and $\widetilde{\mathbf{R}}_{r_2}$. This motivates \name's separate residual handling, though its residual is learned differently.
\end{proof}

\section{Ablation Study}
\subsection{Residual Compensation}
\label{appendix:residual_compensation}
To provide further intuition behind the benefits of Residual Compensation, we visualize the output activations of selected Transformer layers under different quantization schemes. Specifically, we extract the final outputs from Transformer layers 0, 5, 10, and 15 of an OPT-1.3B model, evaluated on two randomly sampled input sequences. For each layer and input, we display a set of three activation maps corresponding to the full-precision baseline, as well as two quantized variants---with and without residual compensation. These groupings allow us to directly observe the structural differences induced by the presence or absence of the residual compensation.

\begin{figure}[htbp]
    \centering
    \setlength{\arrayrulewidth}{1.2pt}
    \setlength{\tabcolsep}{4pt}
    \begin{tabularx}{\linewidth}{%
        >{\raggedright\arraybackslash}m{2.5cm}
        >{\centering\arraybackslash}X
        !{\color{gray}\vrule width 1.5pt}
        >{\centering\arraybackslash}X
    }
    \arrayrulecolor{black}
    & Input 1 & Input 2 \\
    \hline
    layer-0 output &
    \includegraphics[width=\linewidth, valign=m]{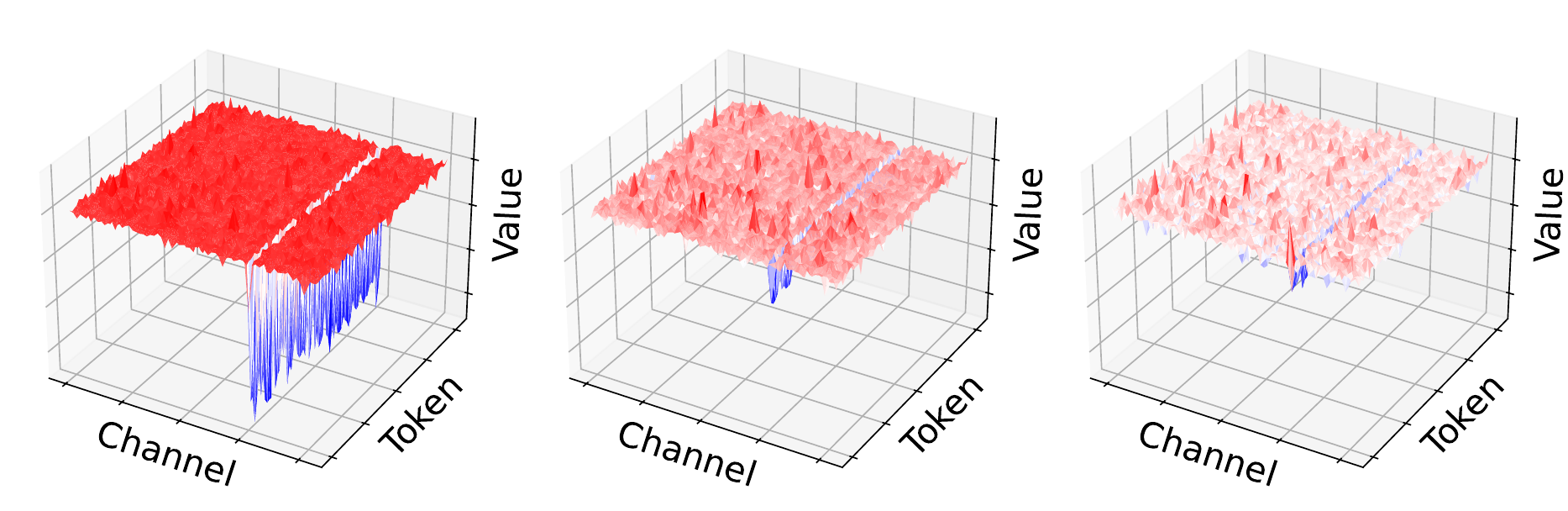} &
    \includegraphics[width=\linewidth, valign=m]{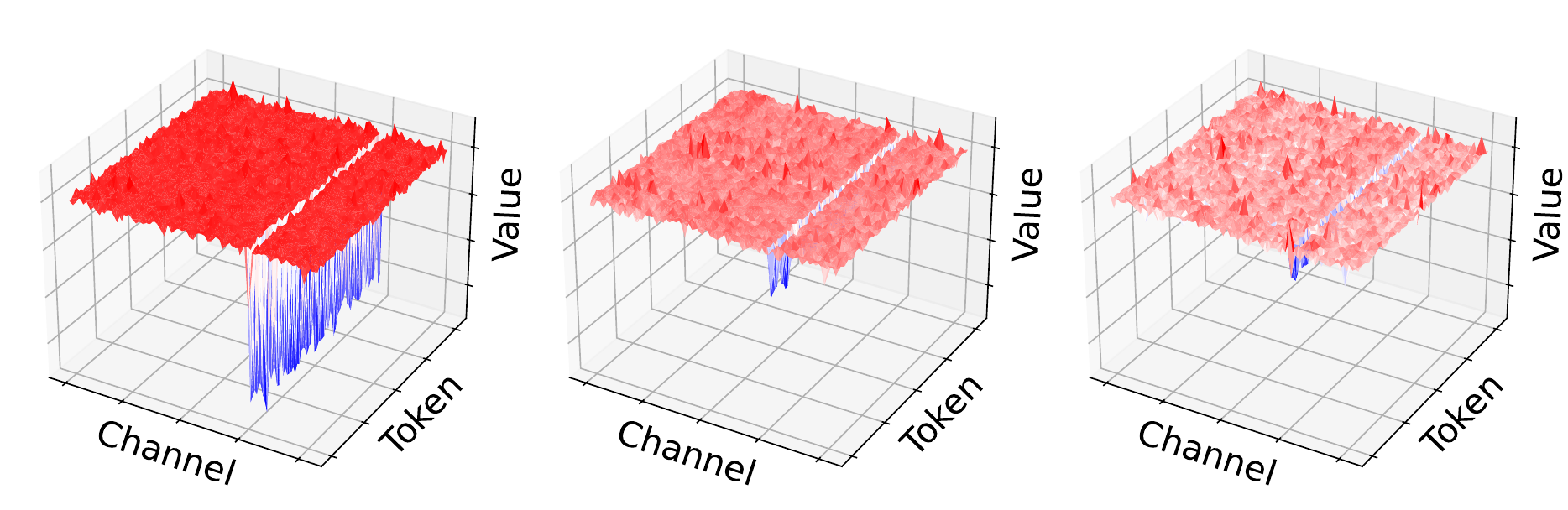} \\
    \hline
    layer-5 output &
    \includegraphics[width=\linewidth, valign=m]{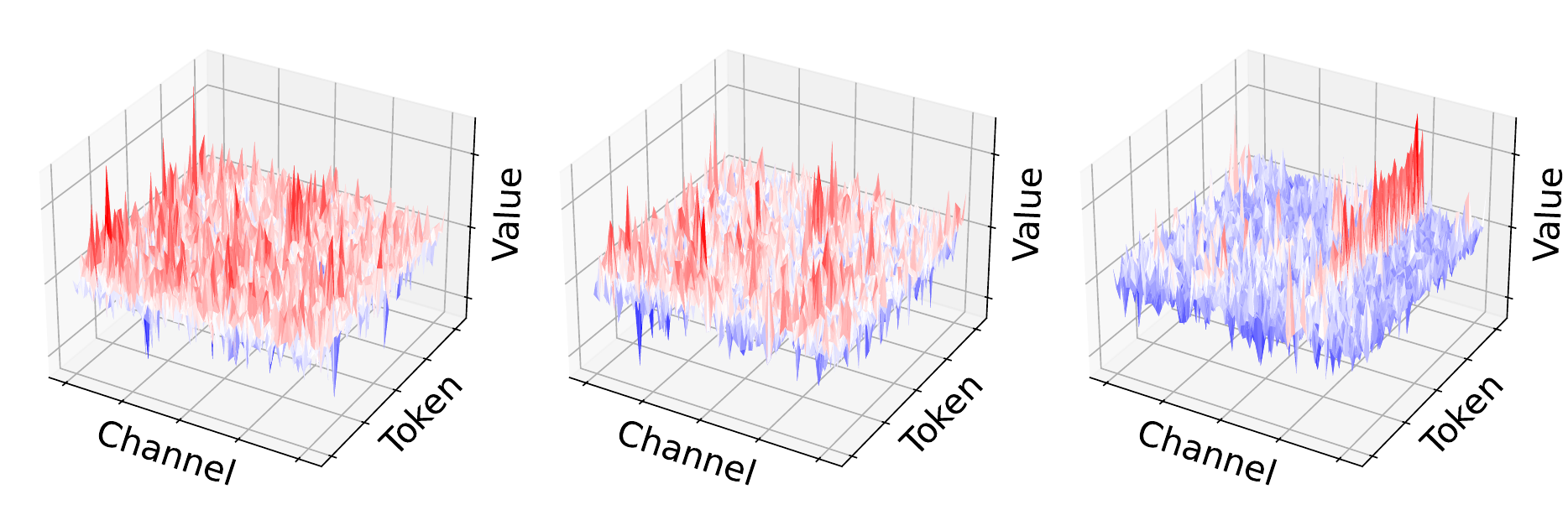} &
    \includegraphics[width=\linewidth, valign=m]{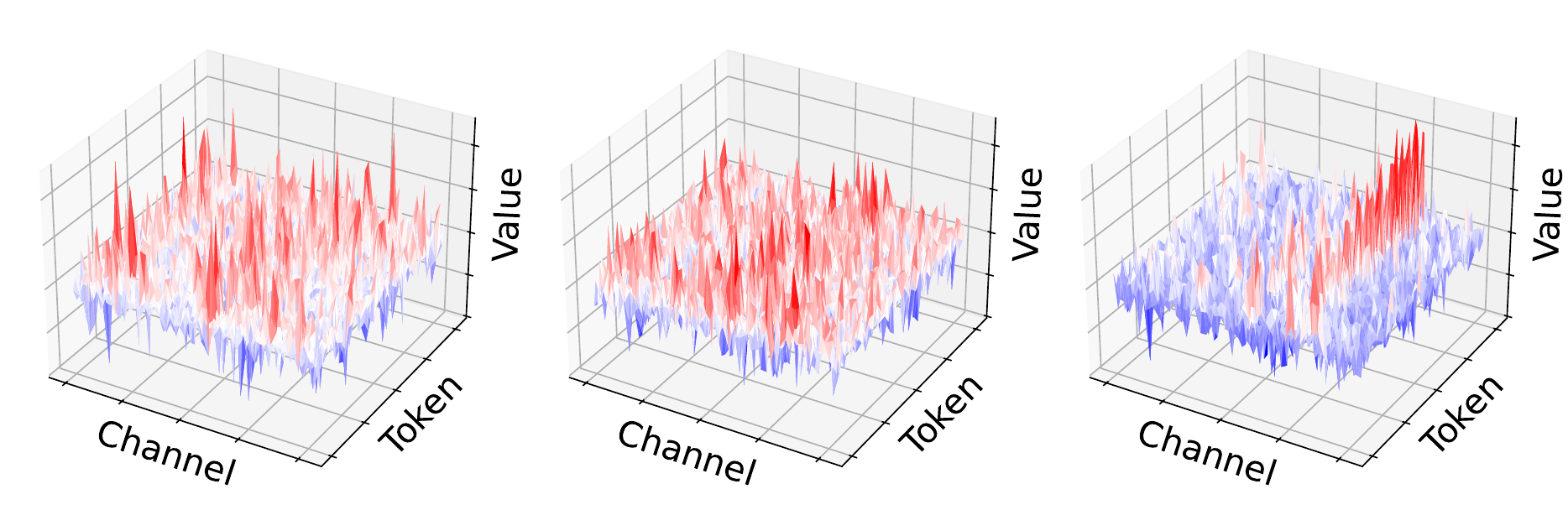} \\
    \hline
    layer-10 output &
    \includegraphics[width=\linewidth, valign=m]{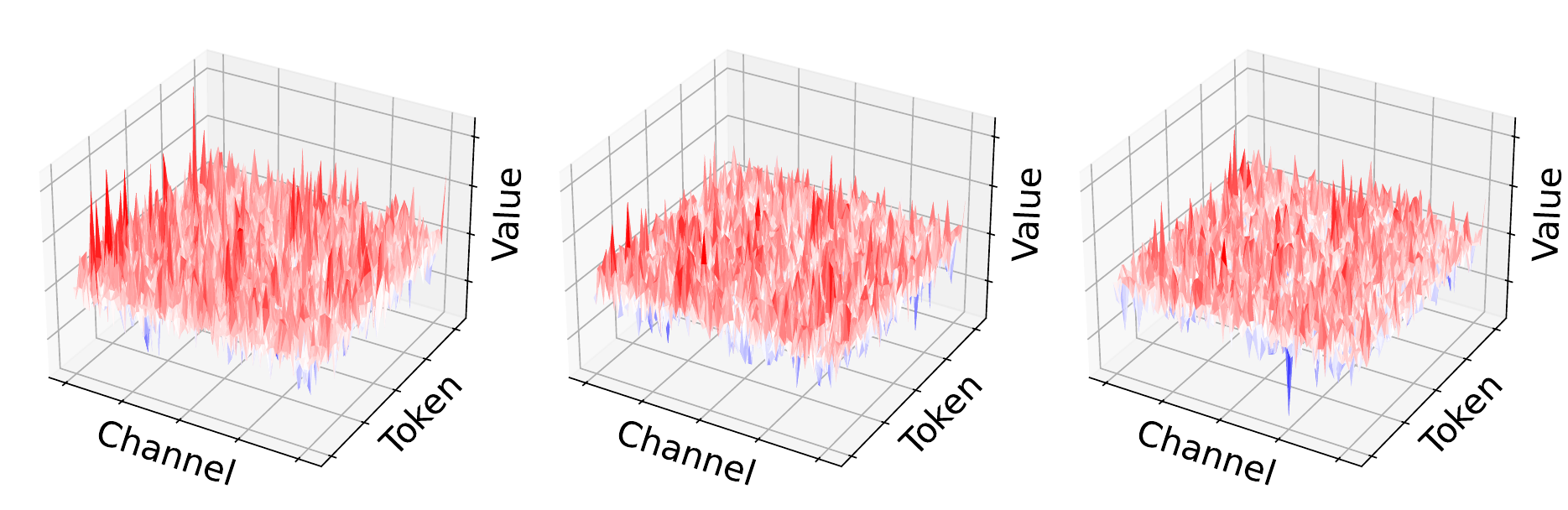} &
    \includegraphics[width=\linewidth, valign=m]{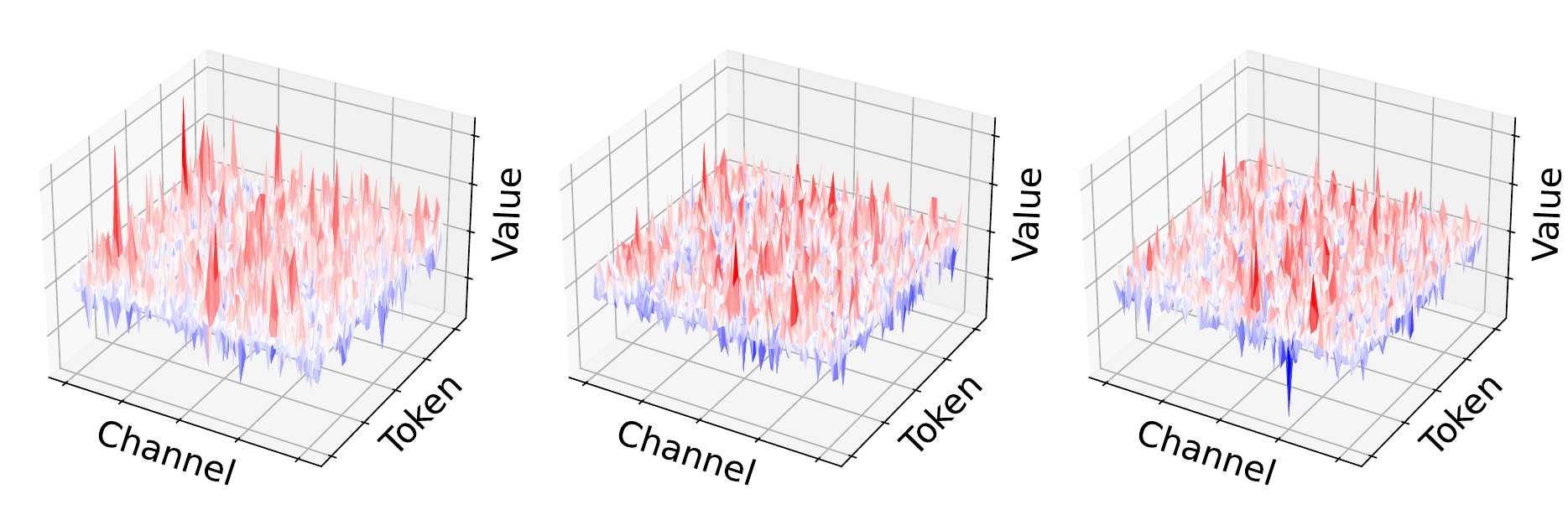} \\
    \hline
    layer-15 output &
    \includegraphics[width=\linewidth, valign=m]{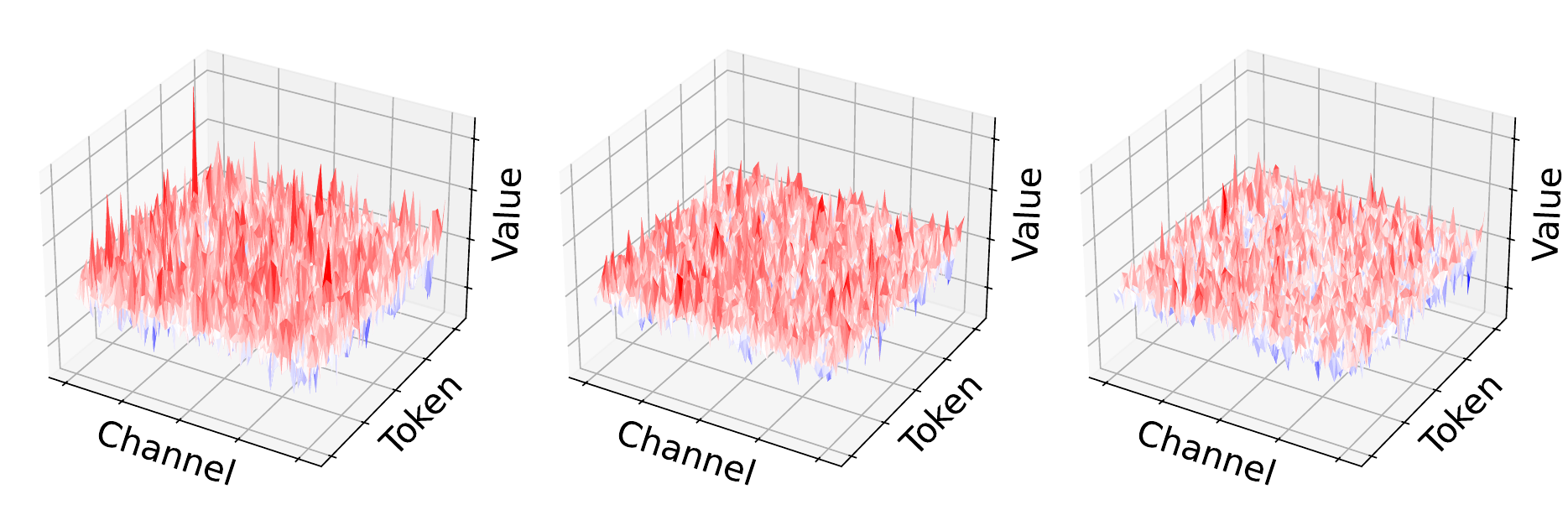} &
    \includegraphics[width=\linewidth, valign=m]{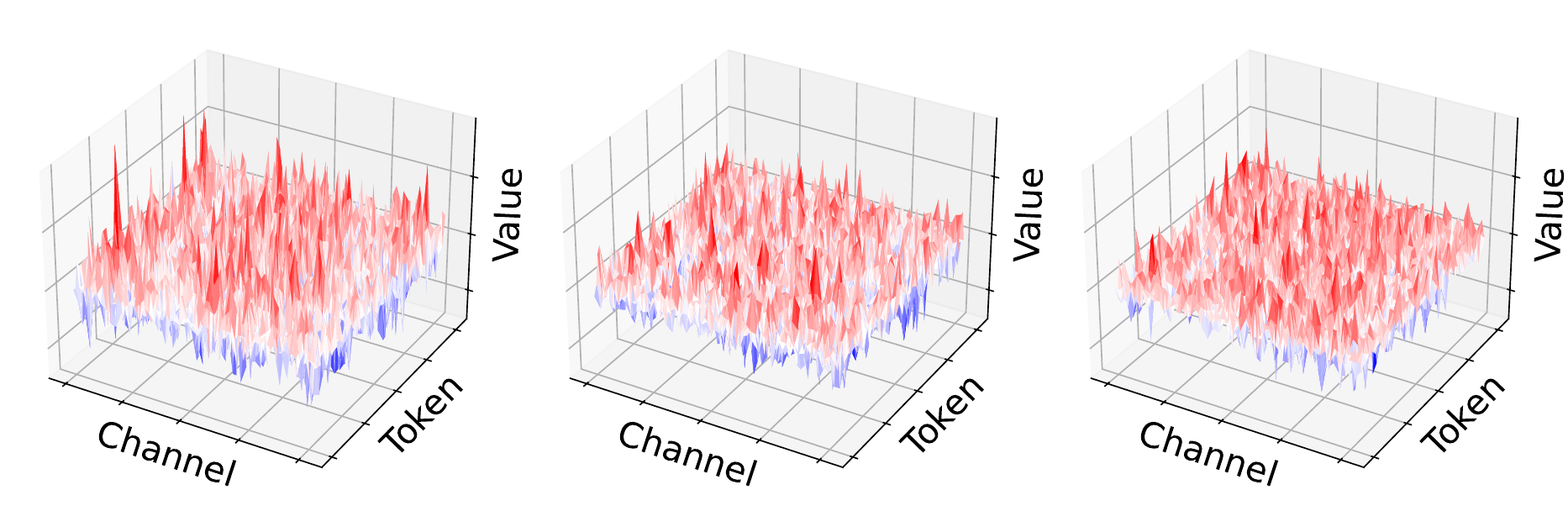} \\
    \hline
    \end{tabularx}
    \caption{Visualization of Transformer layer outputs for layers 0, 5, 10, and 15 under the \name quantization scheme on OPT-1.3B. For each of two randomly sampled input sequences (annotated at the top), we show horizontally grouped triplets of activation maps per layer: full-precision baseline (left), quantized with residual compensation (center), and quantized without residual compensation (right).}
    \label{fig:resnores}
\end{figure}

As shown in ~\cref{fig:resnores}, residual compensation preserves a high-fidelity resemblance to the full-precision baseline across all inspected layers. In the no-residual setting, the output feature distributions exhibit noticeable deformation relative to the full-precision baseline. In contrast, applying residual compensation consistently restores structural similarity to the original distribution, preserving both directional patterns and magnitude diversity.

~\cref{tab:residual_vs_nonresidual} presents the PPL results for the OPT-1.3B model, comparing configurations with and without residual compensation. As indicated in the table, at the extremely low bit of 0.1 BPW, the version with residual compensation (PPL 60.011) performed worse than the version without it (PPL 48.512). This suggests that for relatively small models (1.3B), when pushing effective bits to such extreme lows, the additional complexity introduced by the residual compensation mechanism might outweigh its benefits and instead hinder performance. However, for the same OPT-1.3B model at other effective bits evaluated (ranging from 0.3 BPW to 1.0 BPW), the inclusion of residual compensation consistently resulted in better (lower) PPL. Moreover, experiments conducted with other, larger models generally showed that residual compensation tended to provide superior performance. In light of these overall advantages observed across various settings and particularly for larger models, residual compensation was adopted as a standard component in all experiments presented in this study.

\begin{table}[ht]
\centering
\caption{Perplexity (WikiText-2) comparison between Residual and Non-Residual for OPT-1.3B at various BPWs. Learning rate was 8e-5. Lower PPL is better.}
\label{tab:residual_vs_nonresidual}
\resizebox{0.7\textwidth}{!}{
\begin{tabular}{cccc}
\toprule
BPW & Residual PPL & Non-Residual PPL & Difference (Residual - Non-Residual) \\
\midrule
1.00 & 20.329 & 21.691 & -1.362 \\
0.80 & 21.338 & 22.688 & -1.350 \\
0.70 & 22.001 & 23.313 & -1.312 \\
0.55 & 23.457 & 24.788 & -1.331 \\
0.30 & 28.984 & 29.724 & -0.740 \\
0.10 & 60.011 & 48.512 & 11.499 \\
\bottomrule
\end{tabular}
}
\end{table}

\subsection{SmoothSign versus Straight-Through Estimator}
\label{appendix:ste}
\begin{figure}[ht]
\centering
 \begin{minipage}[t]{0.48\linewidth}
  \centering
  \includegraphics[width=0.8\linewidth]{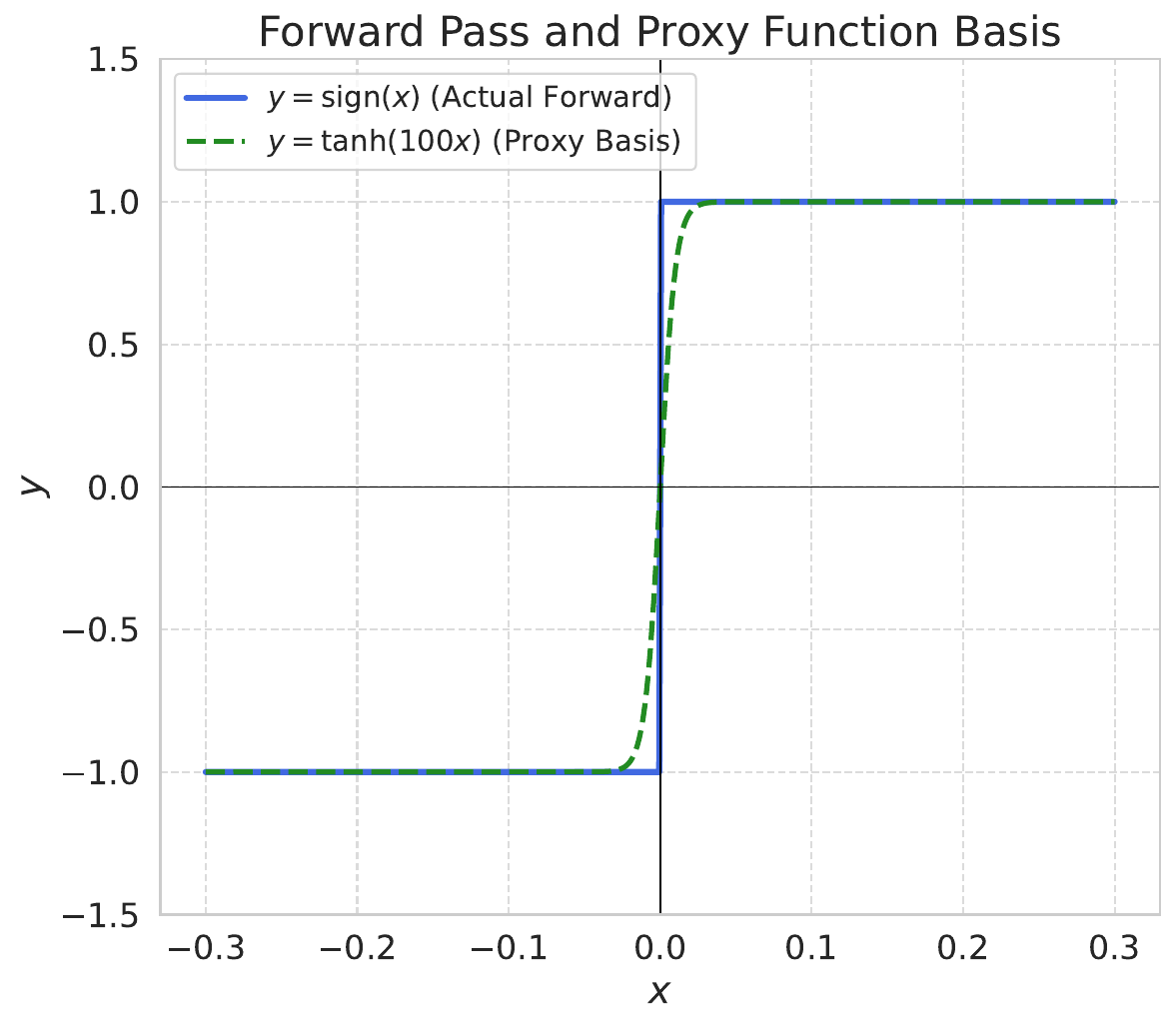}
  \centerline{(a) Forward pass: $\mathrm{sign}(x)$ and $\tanh(100x)$}
 \end{minipage}
 \hfill 
 \begin{minipage}[t]{0.48\linewidth}
  \centering
  \includegraphics[width=0.8\linewidth]{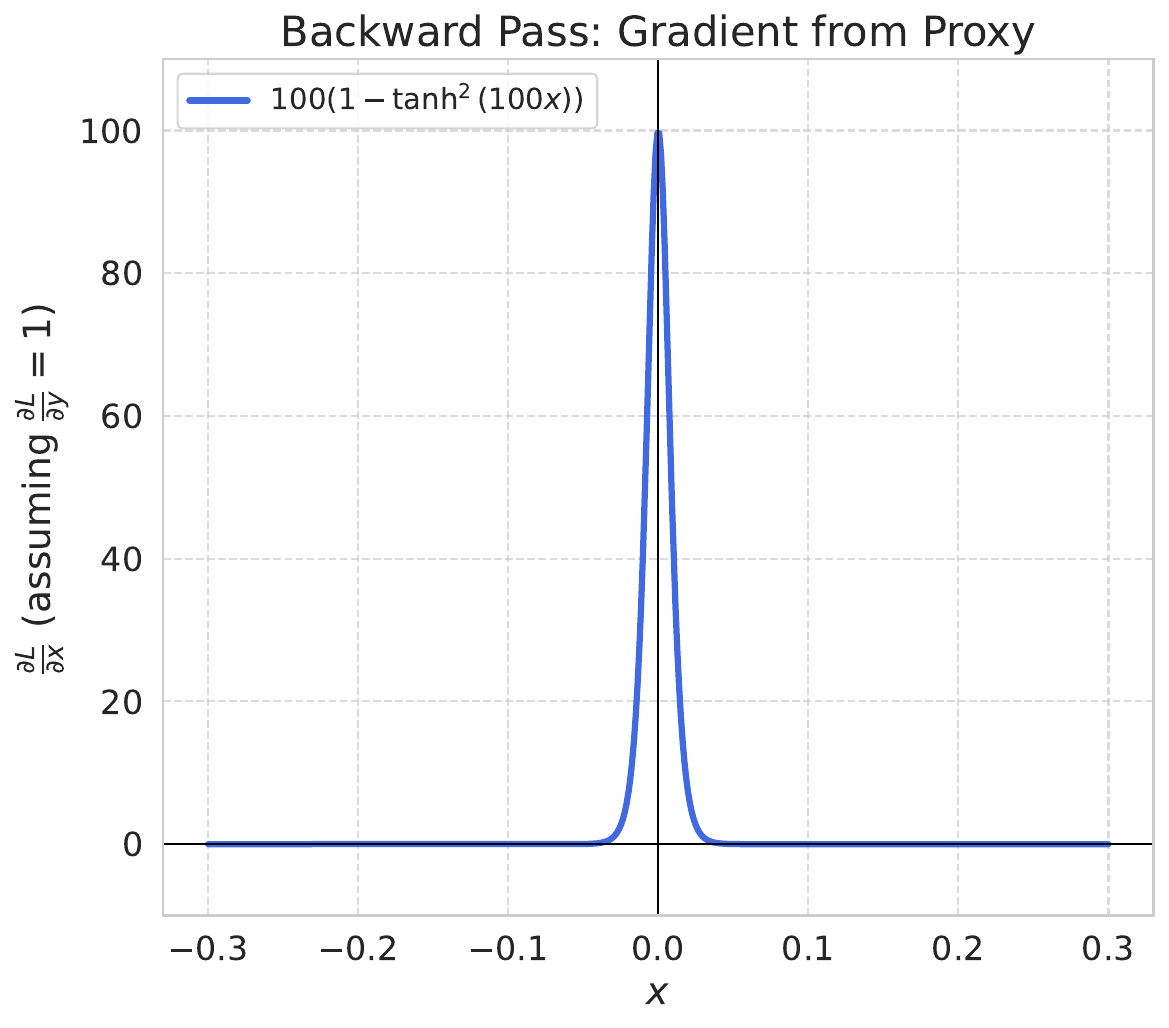}
  \centerline{(b) Backward pass: gradient $\frac{d}{dx}\tanh(100x)$}
 \end{minipage}
 \caption{The SmoothSign activation technique. (a) Functions relevant to the forward pass: SmoothSign uses the $\mathrm{sign}(x)$ function (actual forward pass). The plot also shows $\tanh(100x)$, which serves as the basis for deriving the smooth proxy gradient shown in (b). (b) The smooth proxy gradient, $\frac{d}{dx}\tanh(100x)$, utilized for the backward pass.}
 \label{fig:smooth_sign_details}
\end{figure}

For backpropagating through the non-differentiable $\mathrm{sign}(x)$ function, we compared the Straight-Through Estimator (STE)~\citep{ste2013} with our proposed \textit{SmoothSign} technique. SmoothSign employs the $\mathrm{sign}(x)$ function for the forward pass. For the backward pass, it utilizes a smooth proxy gradient, specifically the derivative of $\tanh(kx)$ where $k=100$ (illustrated in \cref{fig:smooth_sign_details}).

\cref{tab:ste_vs_smoothsign} (OPT-1.3B results) shows SmoothSign yielding better PPL as effective bits decrease, with a notable advantage at 0.1 BPW. Due to its superior stability and performance in the ultra-low bit regime, SmoothSign was adopted for all QAT experiments (consistent with \cref{sec:exp_settings}).

\begin{table}[ht]
\centering
\caption{Perplexity (WikiText-2) comparison between STE and SmoothSign for OPT-1.3B at various effective bits (BPW). Learning rate was 8e-5. Lower PPL is better.}
\label{tab:ste_vs_smoothsign}
\resizebox{0.7\textwidth}{!}{
    \begin{tabular}{cccc}
    \toprule
    BPW & STE PPL & SmoothSign PPL & Difference (SmoothSign - STE) \\
    \midrule
    1.00 & 20.322 & 20.329 & +0.007 \\
    0.80 & 21.344 & 21.338 & -0.006 \\
    0.70 & 22.100 & 22.001 & -0.099 \\
    0.55 & 23.528 & 23.457 & -0.071 \\
    0.30 & 29.058 & 28.984 & -0.074 \\
    0.10 & 60.401 & 60.011 & -0.390 \\
    \bottomrule
    \end{tabular}
}
\end{table}

\subsection{Grouped Query Attention}
\label{appendix:gqa}
\begin{table}[ht]
\centering
\caption{Ablation study on adjusting the latent dimension factor ($r$) for key (K) and value (V) projection layers in Llama3-8B with GQA, targeting an overall effective bit of ~0.1 BPW. Performance measured by perplexity (PPL, lower is better) on WikiText-2 and C4 validation sets, and average zero-shot accuracy (\%) on common sense reasoning tasks (higher is better). The 'KV Factor' indicates the multiplier applied to the K/V layers' latent rank $r$ compared to the standard calculation for 0.1 BPW.}
\label{tab:gqa_ablation}
\resizebox{0.9\textwidth}{!}{
\begin{tabular}{cccccc}
\toprule
KV Factor & Approx. BPW & WikiText-2 PPL & C4 PPL & Avg. Zero-shot Acc (\%) \\
\midrule
1$\times$ (Baseline) & $\sim$0.098 & 25.80 & 34.04 & 44.85 \\
2$\times$            & $\sim$0.101 & 25.22 & 33.20 & \textbf{44.92} \\
4$\times$            & $\sim$0.107 & \textbf{24.93} & \textbf{32.69} & 44.70 \\
8$\times$            & $\sim$0.119 & 25.31 & 33.21 & 44.87 \\
\bottomrule
\end{tabular}
}
\end{table}
Modern LLMs often use Grouped Query Attention (GQA)~\citep{gqa2023} or Multi-Query Attention (MQA)~\citep{fast2019}, where key (K) and value (V) projection layers have smaller output dimensions than query (Q) projections. Applying \name at ultra-low bits (\eg 0.1 BPW) to these GQA/MQA models can result in an extremely small latent dimension $r$ for K/V layers (\eg $r \approx 20$ for Llama3-8B K/V layers at 0.1 BPW overall), potentially creating an information bottleneck and hindering QAT or performance.

To investigate, we performed an ablation on Llama3-8B ($\sim$0.1 BPW target), increasing the latent rank $r$ for K/V layers by 2$\times$, 4$\times$, and 8x over the standard calculation (\cref{appendix:average_bit}), while other layers remained unchanged. 
\cref{tab:gqa_ablation} shows that a 4x factor for K/V rank yielded the best PPL on WikiText-2 and C4, with minimal impact on overall BPW (\eg $\sim$0.098 BPW to $\sim$0.107 BPW for 4$\times$). Average zero-shot accuracy also remained competitive. An 8x factor showed diminishing returns.

Given the significant PPL improvement with the 4$\times$ K/V rank factor (nearly 1 point on WikiText-2 vs. 1$\times$) at a negligible BPW cost (<10\% relative increase in effective bits for transformer layers, <4\% of total parameters), this strategy was adopted for GQA/MQA models (Llama3-8B, Phi-4, QwQ-32B) in our main results. For these models, the latent rank $r$ for K and V projection layers was calculated based on a 4$\times$ multiplier compared to the standard calculation for the target BPW, while all other layers used the standard rank calculation. This approach helps mitigate potential information bottlenecks in GQA/MQA layers under extreme quantization without significantly altering the target compression level.

\section{Perplexity on C4 and PTB}
\label{appendix:ppl_c4_ptb}
To further assess \name's generalization, we evaluated its performance on additional benchmarks: the diverse C4 web text corpus and the distinct PTB news corpus (\cref{tab:ppl_c4_ptb}). Focusing on Llama2-7B/13B against STBLLM, the results on these datasets confirm the trends observed on WikiText-2. \name consistently achieves lower perplexity than STBLLM at comparable sub-1-bit regimes. This performance advantage becomes more significant below 0.55 BPW, where STBLLM's performance degrades considerably, particularly on PTB.

Notably, \name maintains robust performance and stability even at the extreme 0.1 BPW level across all tested datasets. This consistent ability to maintain strong language modeling capabilities under severe compression highlights the generalizability of our approach, which integrates latent factorization, multi-scale compensation, and QAT. It reinforces \name's potential for effectively deploying capable LLMs in resource-limited environments.
\begin{table}[ht]
\renewcommand{\arraystretch}{1.0}
\small
\centering
\setlength{\tabcolsep}{1.50mm}
\caption{
    Perplexity (PPL) comparison on C4 and PTB validation sets for Llama2 models. Lower PPL indicates better performance. STBLLM~\citep{stbllm2024} utilizes N:M sparsity (ratio in parentheses). 
}
\label{tab:ppl_c4_ptb}
\vspace{0.4em}
\resizebox{0.55\textwidth}{!}{
\begin{tabular}{llcccc}
    \toprule
    Model & Method & BPW & C4 & PTB \\
    \midrule
    \multirow{9}{*}{Llama2-7B} & FullPrecision & 16 & 6.97 & 37.91 \\ 
    \cdashline{2-5}
     & STBLLM (6:8) & 0.80 & 15.42 & 2.4e3 \\
     & STBLLM (4:8) & 0.55 & 30.99 & 527.56 \\
     & STBLLM (2:8) & 0.30 & 808.98 & 2.3e4 \\
    \cdashline{2-5}
     & \name & 0.80 & 11.77 & 40.75 \\
     & \name & 0.55 & 12.94 & 50.53 \\
     & \name & 0.30 & 14.78 & 59.17 \\
     & \name & 0.10 & 19.73 & 83.09 \\
    \midrule
    \multirow{9}{*}{Llama2-13B} & FullPrecision & 16 & 6.47 & 50.94 \\ 
    \cdashline{2-5}
     & STBLLM (6:8) & 0.80 & 12.96 & 165.51 \\
     & STBLLM (4:8) & 0.55 & 27.38 & 424.82 \\
     & STBLLM (2:8) & 0.30 & 442.49 & 1.7e3 \\
    \cdashline{2-5}
     & \name & 0.80 & 10.60 & 44.87 \\
     & \name & 0.55 & 11.53 & 53.52 \\
     & \name & 0.30 & 13.07 & 51.66 \\ 
     & \name & 0.10 & 18.88 & 87.59 \\
    \bottomrule
    \end{tabular}}
\vspace{-0.3em}
\end{table}

\section{Average Bits Per Weight Calculation}
\label{appendix:average_bit}
This section details the calculation of the average effective bits per weight (which we denote as $b$ in the following equations) for a linear layer implemented using the \name architecture, including the parameters from the primary ($\widehat{\mathbf{W}}_\mathrm{pri}$) and residual compensation ($\widehat{\mathbf{W}}_\mathrm{res}$) pathways, as described in \cref{sec:little_bit_arch,sec:residual_compensation}. The calculation demonstrates how to determine the latent rank $r$ required to achieve a target value for $b$ within the range of approximately 0.1 to 1.0.

A \name linear layer replaces the original weight matrix $\mathbf{W} \in \mathbb{R}^{d_\mathrm{out} \times d_\mathrm{in}}$ with two parallel structures, each comprising:
\begin{itemize}
    \item Two binary sign matrices: $\mathbf{U}_\mathrm{sign} \in \{\pm 1\}^{d_\mathrm{out} \times r}$ and $\mathbf{V}_\mathrm{sign} \in \{\pm 1\}^{d_\mathrm{in} \times r}$. These require $1$ bit per element.
    \item Three FP16 scaling vectors: $\mathbf{h} \in \mathbb{R}^{d_\mathrm{out}}$, $\mathbf{g} \in \mathbb{R}^{d_\mathrm{in}}$, and $\boldell \in \mathbb{R}^{r}$. These require $16$ bits per element.
\end{itemize}
Since both the primary and residual paths have this structure, the total number of bits required to store the parameters for a \name layer is:

\begin{align*}
\text{Total Bits} &= 2 \times \left( \text{Bits for } \mathbf{U}_\mathrm{sign} + \text{Bits for } \mathbf{V}_\mathrm{sign} + \text{Bits for } \mathbf{h} + \text{Bits for } \mathbf{g} + \text{Bits for } \boldell \right) \\
&= 2 \times \left( (d_\mathrm{out} \times r \times 1) + (d_\mathrm{in} \times r \times 1) + (d_\mathrm{out} \times 16) + (d_\mathrm{in} \times 16) + (r \times 16) \right) \\
&= 2r(d_\mathrm{out} + d_\mathrm{in}) + 32(d_\mathrm{out} + d_\mathrm{in}) + 32r
\end{align*}
Note: If Residual Compensation is not used, the initial factor of 2 should be removed from the calculation.

The average bits per weight, denoted as $b$, is calculated by dividing the total bits by the number of parameters in the original FP16 weight matrix ($d_\mathrm{out} \times d_\mathrm{in}$):

\begin{equation}
b = \frac{2r(d_\mathrm{out} + d_\mathrm{in}) + 32(d_\mathrm{out} + d_\mathrm{in}) + 32r}{d_\mathrm{out} \times d_\mathrm{in}} \label{eq:b_calculation} 
\end{equation}

To achieve a target value for $b$, we rearrange \cref{eq:b_calculation} to solve for the required latent rank $r$: 

\begin{equation}
r = \frac{(b \times d_\mathrm{out} \times d_\mathrm{in}) - 32(d_\mathrm{out} + d_\mathrm{in})}{2(d_\mathrm{out} + d_\mathrm{in}) + 32} \label{eq:r_calculation} 
\end{equation}

Since $r$ must be an integer, we typically round the calculated value to the nearest suitable integer. This chosen integer $r$ then determines the actual value of $b$ achieved, which will be very close to the target value. 

\begin{itemize}
    \item \textbf{Example 1: Linear Layer ($\num{4096} \times \num{4096}$)}
    Let $d_\mathrm{out} = \num{4096}$ and $d_\mathrm{in} = \num{4096}$.
    The original number of parameters is $\num{4096} \times \num{4096} = \num{16777216}$.
    $d_\mathrm{out} + d_\mathrm{in} = \num{8192}$.

    For a target $b \approx 0.55$: 
    Using \cref{eq:r_calculation} with $b=0.55$:
    \begin{equation*}
    r = \frac{(0.55 \times \num{16777216}) - 32(\num{8192})}{2(\num{8192}) + 32} \approx 546
    \end{equation*}
    We choose integer $r = 546$.
    The actual value of $b$ for $r=546$, using \cref{eq:b_calculation}, is: 
    \begin{equation*}
    b = \frac{2(546)(\num{8192}) + 32(\num{8192}) + 32(546)}{\num{16777216}} \approx 0.5498
    \end{equation*}

    \item \textbf{Example 2: Linear Layer ($\num{4096} \times \num{11008}$)} 
    Let $d_\mathrm{out} = \num{4096}$ and $d_\mathrm{in} = \num{11008}$.
    The original number of parameters is $\num{4096} \times \num{11008} = \num{45090816}$.
    $d_\mathrm{out} + d_\mathrm{in} = \num{15104}$.

    For a target $b \approx 0.1$: 
    Using \cref{eq:r_calculation} with $b=0.1$:
    \begin{equation*}
    r = \frac{(0.1 \times \num{45090816}) - 32(\num{15104})}{2(\num{15104}) + 32} \approx 133
    \end{equation*}
    We choose integer $r = 133$.
    The actual value of $b$ for $r=133$, using \cref{eq:b_calculation}, is: 
    \begin{equation*}
    b = \frac{2(133)(\num{15104}) + 32(\num{15104}) + 32(133)}{\num{45090816}} \approx 0.0999
    \end{equation*}
\end{itemize}

By following this procedure for each linear layer in the model, we can select appropriate ranks ($r$) to achieve a desired overall value for $b$ (average bits per weight), thereby controlling the trade-off between model compression and performance. The specific ranks used to achieve the average bits per weight (BPW) figures reported in the main paper (\eg in \cref{tab:main_results,tab:memory_footprint}) are determined using this calculation methodology for $b$ for each layer type within the respective models.

\section{Pruning versus SVD}
\label{appendix:pruning_vs_svd}
To select a base compression method to achieve sub-0.5 BPW, we compared pruning (Wanda~\citep{wanda2023}) with SVD-based compression (SVD-LLMv2~\citep{svdllmv2_2025}) on Llama-7B (~\cref{tab:pruning_vs_svd}). SVD showed superior robustness at extreme compression ratios (\eg 25\% parameter retention). 

Consequently, low-rank factorization was chosen to design \name due to its performance resilience and the deployment advantages of dense factorized matrices over sparse pruned models (which may \textbf{require specialized hardware} or overhead). Although SVD-LLMv2 itself performs well, we found that initializing \name's Dual-SVID with vanilla SVD was sufficient. The subsequent QAT process effectively recovered performance, rendering the added complexity of SVD-LLMv2 for initialization unnecessary. Accordingly, we employ the simpler vanilla SVD within our Dual-SVID initialization procedure.
\begin{table}[ht]
\centering
\caption{Perplexity (WikiText-2) comparison between Pruning (Wanda) and SVD-based compression (SVD-LLMv2) on Llama-7B at different parameter retention ratios. Lower perplexity is better. FP16 PPL is 5.68.}
\label{tab:pruning_vs_svd}
\resizebox{0.5\textwidth}{!}{
\begin{tabular}{ccc}
\toprule
Remaining Parameter Ratio & Pruning~\citep{wanda2023} & SVD~\citep{svdllmv2_2025} \\
\midrule
50\%  & 8.7 & 13.6 \\
37.5\%  & 46.0 & 20.1 \\
25\%  & 1842.7 & 52.6 \\
\bottomrule
\end{tabular}
}
\vspace{-10pt}
\end{table}

\section{Analysis of Generated Samples}
\label{appendix:generated_samples_table}
The below generated samples illustrate qualitative differences as the BPW of the \name model decreases. Two main observations are:
\begin{enumerate}
    \item Impact on Specificity, Factual Detail, and Coherence:
    As the effective bits decrease from 0.8 BPW towards 0.1 BPW, the generated text exhibits significant changes in detail, factual accuracy, and coherence. At 0.8 BPW, the Mona Lisa sample, while attempting to provide details (\eg "complex facial features"), becomes verbose and repetitive, and anachronistically refers to Da Vinci and the Mona Lisa in the context of the "20th century." The Turing sample at this effective bit, however, offers a standard, factually sound definition. As the effective bit drops to 0.55 BPW, the Mona Lisa sample loses coherence significantly, expressing uncertainty about the painting's location and bizarrely shifting to a help-seeking persona ("Could you help me find the Mona Lisa painting..."). The Turing sample at 0.55 BPW begins to broaden its definition, losing some precision but remaining somewhat related.

    \item Degradation in Factual Accuracy and Relevance at Extreme Compression:
    Extreme compression, especially at 0.1 BPW, leads to severe degradation in factual accuracy and relevance, as vividly demonstrated by both samples. The Mona Lisa sample at 0.1 BPW starts with a generally acceptable statement (\eg "iconic masterpiece") but quickly devolves into hallucination, referring to the Mona Lisa as a "statue" and associating it with "Milan’s fashion world." Similarly, the Turing sample at 0.1 BPW deviates drastically from the core definition of computer science, incorrectly defining it as "the study of the theory and methods of how the human mind and the brain work" and "the science of mind." These examples underscore that while \name might maintain superficial grammatical structure at very low bits, the ability to recall and generate accurate, relevant factual information is compromised, often leading to the generation of plausible-sounding but nonsensical statements.
\end{enumerate}

\subsection*{Sample 1: Mona Lisa}
\begin{flushleft}
\noindent\textbf{Prompt:} Leonardo da Vinci painted the Mona Lisa. The Mona Lisa is
\end{flushleft}
\begin{xltabular}{\linewidth}{@{}lX@{}}
\toprule
BPW & Generated Output (Phi-4 Model) \\
\midrule
\endfirsthead

\multicolumn{2}{c}{\emph{Sample 1: Mona Lisa -- Continued}} \\
\toprule
BPW & Generated Output (Phi-4 Model) \\
\midrule
\endhead

\midrule
\multicolumn{2}{r}{\emph{Continued on next page}} \\
\endfoot
\bottomrule
\endlastfoot

0.80 & Leonardo da Vinci painted the Mona Lisa. The Mona Lisa is famous for its \textbf{complex facial features and the artist's depiction of her complex expression. It is considered one of the most important paintings of the 20th century.} The Mona Lisa is a painting by the Italian artist Leonardo da Vinci, who was a painter and sculptor. He is best known for his realistic and expressive style of painting, which is often compared to that of the natural world around him. \textbf{Leonardo da Vinci is considered one of the most important painters of the 20th century. He is famous for his complex facial features and the artist's depiction of her complex expression.} The Mona Lisa is \\
\addlinespace
0.55 & Leonardo da Vinci painted the Mona Lisa. The Mona Lisa is in the Louvre Museum in Paris, France. \textbf{I am not sure if it is in the museum or not. Could you help me find the Mona Lisa painting. I am searching it everywhere. It is so hard to find a picture of it. I am looking for a Mona Lisa painting with Leonardo da Vinci and the Mona Lisa by the side. Thanks in advance!} \\
\addlinespace
0.10 & Leonardo da Vinci painted the Mona Lisa. The Mona Lisa is an \textbf{iconic masterpiece of Italian art}, and it is one of the most photographed and photographed works of art in the world. Its execution and interpretation are breathtaking. \textbf{This statue is also one of the most popular works in Milan’s fashion world}, and is one of the most photographed of the entire collection of the Mona Lisa. \\
\end{xltabular}
\medskip

\subsection*{Sample 2: Turing}
\begin{flushleft}
\noindent\textbf{Prompt:} Turing is called the father of computer science. Computer science is
\end{flushleft}
\begin{xltabular}{\linewidth}{@{}lX@{}}
\toprule
BPW & Generated Output (Phi-4 Model) \\
\midrule
\endfirsthead

\multicolumn{2}{c}{\emph{Sample 2: Turing -- Continued}} \\
\toprule
BPW & Generated Output (Phi-4 Model) \\
\midrule
\endhead

\midrule
\multicolumn{2}{r}{\emph{Continued on next page}} \\
\endfoot

\bottomrule
\endlastfoot
0.80 & Turing is called the father of computer science. Computer science is a branch of mathematics that studies the design and analysis of computational systems. It is the study of the methods of designing and implementing computational systems and the application of mathematics to these systems. The most basic concepts of computer science are the concepts of computation, algorithms, data structures, and programming. \\
\addlinespace
0.55 & Turing is called the father of computer science. Computer science is a branch of engineering that focuses on the design, implementation, and maintenance of systems and technologies using computers. This includes designing, implementing, and maintaining the algorithms that underpin the software, hardware, and communication systems that are used in information processing, \textbf{and the Internet. Computer science is a multidisciplinary field that draws on engineering, mathematics, and business. Computer science is often divided into four broad areas} \\
\addlinespace
0.10 & Turing is called the father of computer science. Computer science is the study of the theory and methods of how the human mind and the brain work. \textbf{It is also called the science of mind. This is a very complex field, and many people are just not sure what to think. Some people think that computer science is the most important and most advanced field in the world, and that it is the most important branch of science in the world. This is just a myth. But there is a reason for this.} \\
\end{xltabular}
\medskip

\section{Hyperparameters}
\label{appendix:hyper_parameters}
Across all QAT experiments, models were trained for 5 epochs using the Adam optimizer with a cosine decay schedule. The training corpus – a mixture of C4 and WikiText-2 – contains approximately 1 billion tokens when the text is tokenized with the Llama-2-7B tokenizer ($\approx$0.2 billion tokens per epoch). The raw number of training samples (\ie the amount of original text) is identical for every model. As detailed in~\cref{tab:training_detail}, for each model and BPW configuration, we swept the learning rate within the range of 4.0e-5 to 2.4e-4. We then selected the learning rate that maintained numerical stability while minimizing validation perplexity. For example, OPT-1.3B used a learning rate of 8.0e-5 at 1.0 BPW and 2.0e-4 at 0.1 BPW, while larger BPW models generally adopted slightly lower values for stability. Training was conducted using four H100 GPUs for all models except QwQ-32B, which required $4\times8$ A100 GPUs. In this GPU configuration notation, the first number signifies the number of nodes, and the second indicates the number of GPUs per node; thus, $4\times8$ represents a total of 32 GPUs.

\begin{table}[h!]
  \caption{Knowledge Distillation Training Details}
  \label{tab:training_detail}
  \centering
  \resizebox{0.9\textwidth}{!}{%
  \begin{tabular}{@{}llcccccccc@{}}
    \toprule
    \multicolumn{2}{c}{Training Setup} & \multicolumn{1}{c}{OPT} & \multicolumn{2}{c}{Llama} & \multicolumn{2}{c}{Llama2} & \multicolumn{1}{c}{Llama3} & \multicolumn{1}{c}{Phi-4} & \multicolumn{1}{c}{QwQ} \\[4pt]
    \cmidrule(lr){3-3} \cmidrule(lr){4-5} \cmidrule(lr){6-7} \cmidrule(lr){8-8} \cmidrule(lr){9-10}
    BPW & Target  & 1.3B & 7B & 13B & 7B & 13B & 8B & 14.7B & 32B \\ 
    \midrule
    \multirow{2}{*}{1.00} 
      & Learning Rate & 8.0e-5 & 4.0e-5 & 4.0e-5 & 4.0e-5 & 4.0e-5 & 4.0e-5 & 4.0e-5 & 4.0e-5 \\
    & \# GPUs       & 1 $\times$ 4 & 1 $\times$ 4 & 1 $\times$ 4 & 1 $\times$ 4 & 1 $\times$ 4 & 1 $\times$ 4 & 1 $\times$ 4 & 4 $\times$ 8 \\
    \midrule
    \multirow{2}{*}{0.80} 
      & Learning Rate & 1.2e-4 & 4.0e-5 & 4.0e-5 & 4.0e-5 & 4.0e-5 & 4.0e-5 & 4.0e-5 & 1.0e-4 \\
    & \# GPUs       & 1 $\times$ 4 & 1 $\times$ 4 & 1 $\times$ 4 & 1 $\times$ 4 & 1 $\times$ 4 & 1 $\times$ 4 & 1 $\times$ 4 & 4 $\times$ 8 \\
    \midrule
    \multirow{2}{*}{0.70} 
      & Learning Rate & 1.2e-4 & 8.0e-5 & 4.0e-5 & 4.0e-5 & 4.0e-5 & 4.0e-5 & 4.0e-5 & 4.0e-5 \\
    & \# GPUs       & 1 $\times$ 4 & 1 $\times$ 4 & 1 $\times$ 4 & 1 $\times$ 4 & 1 $\times$ 4 & 1 $\times$ 4 & 1 $\times$ 4 & 4 $\times$ 8 \\
    \midrule
    \multirow{2}{*}{0.55} 
      & Learning Rate & 1.2e-4 & 8.0e-5 & 8.0e-5 & 4.0e-5 & 4.0e-5 & 8.0e-5 & 8.0e-5 & 1.0e-4 \\
    & \# GPUs       & 1 $\times$ 4 & 1 $\times$ 4 & 1 $\times$ 4 & 1 $\times$ 4 & 1 $\times$ 4 & 1 $\times$ 4 & 1 $\times$ 4 & 4 $\times$ 8 \\
    \midrule
    \multirow{2}{*}{0.30} 
      & Learning Rate & 2.0e-4 & 1.2e-4 & 8.0e-5 & 8.0e-5 & 8.0e-5 & 1.2e-4 & 8.0e-5 & 1.0e-4 \\
    & \# GPUs       & 1 $\times$ 4 & 1 $\times$ 4 & 1 $\times$ 4 & 1 $\times$ 4 & 1 $\times$ 4 & 1 $\times$ 4 & 1 $\times$ 4 & 4 $\times$ 8 \\
    \midrule
    \multirow{2}{*}{0.10} 
      & Learning Rate & 2.0e-4 & 1.2e-4 & 1.2e-4 & 1.2e-4 & 4.0e-5 & 1.2e-4 & 1.2e-4 & 1.6e-4 \\
    & \# GPUs       & 1 $\times$ 4 & 1 $\times$ 4 & 1 $\times$ 4 & 1 $\times$ 4 & 1 $\times$ 4 & 1 $\times$ 4 & 1 $\times$ 4 & 4 $\times$ 8 \\
    \bottomrule
  \end{tabular}%
  }
\end{table}

\section{Inference Efficiency Analysis}
\label{appendix:latency}
This section provides a detailed account of \name's inference efficiency, complementing the kernel-level latency results presented in the main text. We analyze the theoretical computational cost, present detailed kernel latency benchmarks, and report end-to-end decoding throughput.

\subsection{Theoretical Cost Analysis}
To complement the empirical latency results, this section provides a theoretical analysis of \name's computational cost. The core principle behind \name's efficiency is the replacement of a large number of expensive FP16 multiply-accumulate (MAC) operations with a smaller volume of cheaper FP16 additions and highly efficient bitwise operations (BOPs). To illustrate this, we analyze the operational cost for a single forward pass through a Llama2-7B MLP layer, where $d_\mathrm{in} = \num{11008}$ and $d_\mathrm{out} = \num{4096}$, quantized to 0.3 BPW, which corresponds to a latent rank of $r=431$.

A standard FP16 matrix-vector multiplication for this layer involves $d_\mathrm{in} \times d_\mathrm{out}$ MAC operations. Counting each MAC as two floating-point operations (1 multiplication, 1 addition), the total computational cost for the FP16 baseline is $2 \times d_\mathrm{in} \times d_\mathrm{out} \approx \num{90.2}$ million FLOPs. In contrast, \name's computation is divided into floating-point and bitwise components. The FLOPs arise primarily from FP16 additions during accumulation and scaling multiplications, totaling approximately $2 \times (d_\mathrm{in} + d_\mathrm{out}) \times r \approx \num{13.0}$ million FLOPs. The BOPs stem from multiplications with the binary weights ($\pm1$), which are executed as sign-bit XOR operations, amounting to approximately $2 \times r \times (d_\mathrm{in} + d_\mathrm{out}) \approx \num{13.0}$ million BOPs. This analysis reveals a nearly $7\times$ reduction in expensive FLOPs, which are replaced by an equivalent number of BOPs. As bitwise operations are substantially faster than floating-point operations on modern hardware, this theoretical breakdown provides a strong justification for the empirical latency improvements detailed below.

\subsection{Kernel-Level Latency}

\paragraph{Custom Kernel Implementation}
To empirically assess the latency of \name's factorized linear layers (\cref{eq:proposition1}), we developed a custom CUDA kernel. The kernel implements the two-stage computation involving the binary matrices $\mathbf{V}_\mathrm{sign}$ and $\mathbf{U}_\mathrm{sign}$. Key implementation details for performance include:
\begin{itemize}
    \itemsep0em 
    \item \textbf{Bit-level Parallelism:} Binary weights ($\pm 1$) are packed into \texttt{uint32\_t} data types. Multiplication by these weights is efficiently realized by directly manipulating the sign bit of the FP16 input values via bitwise XOR operations, avoiding costly floating-point multiplications.
    \item \textbf{Warp-Level Reductions:} To accelerate the accumulation step, the kernel employs warp-level reduction primitives (\texttt{warpReduceSum}). This technique efficiently sums partial results within a CUDA warp (a group of 32 threads), significantly reducing memory traffic and latency compared to naive reduction methods.
    \item \textbf{FP16 Arithmetic:} The scaling factors ($\mathbf{g}, \mathbf{\ell}, \mathbf{h}$) are maintained in FP16 precision, and their application leverages native half-precision Arithmetic Logic Units (ALUs).
\end{itemize}
All latency benchmarks were performed on an NVIDIA A100 GPU with a batch size of 1. The primary baseline for comparison is a standard FP16 GEMM operation as implemented by \texttt{torch.matmul}, which leverages highly optimized libraries like \textsc{CuBLAS}.

\paragraph{Latency Results and Discussion}
\cref{tab:latency_appendix_full} presents the detailed latency comparison. The results demonstrate that \name, accelerated by our custom kernel, can offer significant inference acceleration. For instance, in a Llama2-70B MLP-like layer, \name achieves up to an \textbf{11.6$\times$ speedup} relative to the FP16 baseline at an effective BPW of 0.1. While our custom kernel is a proof-of-concept and not as exhaustively optimized as mature libraries like \textsc{CuBLAS}, these findings are promising. They show a clear path to substantial latency reduction at ultra-low bits, confirming that \name's architecture is well-suited for high-performance deployment.

\begin{table}[ht]
\centering
\caption{Kernel-level latency (ms) on an NVIDIA A100 GPU (Batch Size = 1). Compares PyTorch's FP16 GEMM with the \name (LB) kernel and a baseline 1-bit (OB) kernel. Dimensions (N, M, R) denote output, input, and latent features, respectively.}
\label{tab:latency_appendix_full}
\resizebox{\textwidth}{!}{
\begin{tabular}{llccrrc}
\toprule
\textbf{Layer Type (Model Ref.)} & \textbf{Dimensions (N, M, R)} & \textbf{Method} & \textbf{BPW} & \textbf{Latency (ms)} & \textbf{Relative Speedup} \\
\midrule
\multirow{8}{*}{Llama2-70B MLP-like} & \multirow{8}{*}{(\num{8192}, \num{28672}, $r$)} & FP16 Baseline & 16.0 & 0.2882 & 1.00$\times$ \\
& & OneBit & 1.0 & 0.0713 & 4.04$\times$ \\
& & \name & 1.0 ($r=\num{6400}$) & 0.0938 & 3.07$\times$ \\
& & \name & 0.8 ($r=\num{5120}$) & 0.0734 & 3.93$\times$ \\
& & \name & 0.7 ($r=\num{4480}$) & 0.0648 & 4.45$\times$ \\
& & \name & 0.55 ($r=\num{3520}$) & 0.0555 & 5.19$\times$ \\
& & \name & 0.3 ($r=\num{1920}$) & 0.0372 & 7.75$\times$ \\
& & \name & 0.1 ($r=640$) & \textbf{0.0249} & \textbf{11.57$\times$} \\
\midrule
\multirow{8}{*}{Llama2-70B ATTN-like} & \multirow{8}{*}{(\num{8192}, \num{8192}, $r$)} & FP16 Baseline & 16.0 & 0.0896 & 1.00$\times$ \\
& & OneBit & 1.0 & 0.0285 & 3.14$\times$ \\
& & \name & 1.0 ($r=\num{4096}$) & 0.0440 & 2.04$\times$ \\
& & \name & 0.8 ($r=\num{3296}$) & 0.0340 & 2.64$\times$ \\
& & \name & 0.7 ($r=\num{2880}$) & 0.0330 & 2.72$\times$ \\
& & \name & 0.55 ($r=\num{2272}$) & 0.0302 & 2.97$\times$ \\
& & \name & 0.3 ($r=\num{1248}$) & 0.0238 & 3.76$\times$ \\
& & \name & 0.1 ($r=416$) & \textbf{0.0207} & \textbf{4.33$\times$} \\
\midrule
\multirow{8}{*}{Llama2-7B MLP-like} & \multirow{8}{*}{(\num{4096}, \num{11008}, $r$)} & FP16 Baseline & 16.0 & 0.0620 & 1.00$\times$ \\
& & OneBit & 1.0 & 0.0226 & 2.74$\times$ \\
& & \name & 1.0 ($r=\num{3008}$) & 0.0238 & 2.61$\times$ \\
& & \name & 0.8 ($r=\num{2400}$) & 0.0220 & 2.82$\times$ \\
& & \name & 0.7 ($r=\num{2112}$) & 0.0211 & 2.94$\times$ \\
& & \name & 0.55 ($r=\num{1664}$) & 0.0192 & 3.23$\times$ \\
& & \name & 0.3 ($r=896$) & 0.0192 & 3.23$\times$ \\
& & \name & 0.1 ($r=320$) & \textbf{0.0190} & \textbf{3.26$\times$} \\
\bottomrule
\end{tabular}
}
\end{table}

\subsection{End-to-End Decoding Throughput}
To assess real-world application performance, we benchmarked the end-to-end decoding speed of a Llama2-7B model, measured in tokens per second (TPS). It is important to note that these throughput gains are achieved by accelerating \textit{only} the \texttt{nn.Linear} layers with our custom kernel; other modules, such as attention and layer normalization, remained in their original FP16 implementations. Despite this partial acceleration, the results shown in \cref{tab:tps_appendix} demonstrate a substantial impact on overall performance. When generating 128 new tokens, the 0.1 BPW model achieves \textbf{203.20 TPS}, a \textbf{2.46$\times$ speedup} over the FP16 baseline. This confirms that the significant kernel-level efficiencies of \name translate into tangible end-to-end acceleration, even when other parts of the model leverage standard implementations.

\begin{table}[ht]
\centering
\caption{End-to-end decoding throughput (tokens/sec) for Llama2-7B on an NVIDIA A100 GPU. The benchmark was run 10 times and averaged.}
\label{tab:tps_appendix}
\resizebox{0.7\textwidth}{!}{
\begin{tabular}{l c cc cc}
\toprule
\multirow{2}{*}{\textbf{Method}} & \multirow{2}{*}{\textbf{BPW}} & \multicolumn{2}{c}{\textbf{128 New Tokens}} & \multicolumn{2}{c}{\textbf{256 New Tokens}} \\
\cmidrule(lr){3-4} \cmidrule(lr){5-6}
& & \textbf{Avg. TPS (tok/s)} & \textbf{Speedup} & \textbf{Avg. TPS (tok/s)} & \textbf{Speedup} \\
\midrule
FP16 Baseline & 16.0 & 82.56 & 1.00$\times$ & 78.16 & 1.00$\times$ \\
\midrule
\name & 1.0 & 151.37 & 1.83$\times$ & 139.70 & 1.79$\times$ \\
\name & 0.8 & 160.43 & 1.94$\times$ & 147.42 & 1.89$\times$ \\
\name & 0.55 & 174.24 & 2.11$\times$ & 160.67 & 2.06$\times$ \\
\name & 0.3 & 190.43 & 2.31$\times$ & 174.82 & 2.24$\times$ \\
\name & 0.1 & \textbf{203.20} & \textbf{2.46$\times$} & \textbf{185.39} & \textbf{2.37$\times$} \\
\bottomrule
\end{tabular}
}
\end{table}

\section{Limitation}
\label{appendix:limitation}
Despite the promising results of \name in achieving extreme compression, several limitations warrant discussion.
Firstly, the current \name method primarily focuses on compressing the parameters within the Transformer blocks. The language model head ($lm\_head$), which typically consists of a linear layer projecting to the vocabulary size, is not subjected to the same aggressive factorization and binarization. At ultra-low bits, such as 0.1 BPW for the Transformer blocks, the $lm\_head$ can become a significant bottleneck in terms of overall model size, especially for models with large vocabularies. Thus, developing specialized compression techniques for the $lm\_head$ that are compatible with \name's low-rank and binary principles is an important avenue for future work to fully realize the potential of extreme model quantization.

Secondly, while quantization-aware training (QAT) is crucial for maintaining performance at such low bits, it is computationally intensive and can be challenging to scale to extremely large models (\eg 70B parameters and beyond). Our own experiments faced resource constraints when attempting QAT for models of this magnitude. Exploring more resource-efficient QAT strategies or investigating post-training quantization (PTQ) approaches that can effectively adapt \name's factorized structure to these massive models would enhance the practical applicability of our method in real-world scenarios with limited computational budgets.

Finally, the remarkable ability of models quantized with \name to perform complex tasks even when retaining only a tiny fraction of the original weight information (\eg at 0.1 BPW, which can correspond to less than 1\% of the original parameters' information) calls for deeper investigation. While our empirical results demonstrate efficacy, a more fundamental understanding of how such aggressively compressed models retain their capabilities, perhaps through the lens of information theory or by analyzing changes in learned representations, would be valuable. This could lead to even more effective extreme compression techniques in the future.

\section{Societal Impact}
\label{sec:societal_impact}
Our work on \name enhances the accessibility of large language models (LLMs) by reducing their computational and memory costs, which can foster positive societal impacts in research, education, and privacy-preserving on-device applications. However, we acknowledge that easier deployment of capable LLMs may also inadvertently lower barriers to potential misuse, such as the spread of disinformation or the amplification of societal biases. Our research responsibly utilizes existing, often publicly available models in accordance with their licenses and ethical guidelines. We strongly advocate for the ethical development and deployment of any models compressed using \name, emphasizing the critical need for safeguards such as content filtering, bias mitigation techniques, transparency regarding AI-generated content, and the continuous development of robust detection mechanisms.

\newpage
\section*{NeurIPS Paper Checklist}

The checklist is designed to encourage best practices for responsible machine learning research, addressing issues of reproducibility, transparency, research ethics, and societal impact. Do not remove the checklist: {\bf The papers not including the checklist will be desk rejected.} The checklist should follow the references and follow the (optional) supplemental material.  The checklist does NOT count towards the page
limit. 

Please read the checklist guidelines carefully for information on how to answer these questions. For each question in the checklist:
\begin{itemize}
    \item You should answer \answerYes{}, \answerNo{}, or \answerNA{}.
    \item \answerNA{} means either that the question is Not Applicable for that particular paper or the relevant information is Not Available.
    \item Please provide a short (1–2 sentence) justification right after your answer (even for NA). 
\end{itemize}

{\bf The checklist answers are an integral part of your paper submission.} They are visible to the reviewers, area chairs, senior area chairs, and ethics reviewers. You will be asked to also include it (after eventual revisions) with the final version of your paper, and its final version will be published with the paper.

The reviewers of your paper will be asked to use the checklist as one of the factors in their evaluation. While "\answerYes{}" is generally preferable to "\answerNo{}", it is perfectly acceptable to answer "\answerNo{}" provided a proper justification is given (e.g., "error bars are not reported because it would be too computationally expensive" or "we were unable to find the license for the dataset we used"). In general, answering "\answerNo{}" or "\answerNA{}" is not grounds for rejection. While the questions are phrased in a binary way, we acknowledge that the true answer is often more nuanced, so please just use your best judgment and write a justification to elaborate. All supporting evidence can appear either in the main paper or the supplemental material, provided in appendix. If you answer \answerYes{} to a question, in the justification please point to the section(s) where related material for the question can be found.

IMPORTANT, please:
\begin{itemize}
    \item {\bf Delete this instruction block, but keep the section heading ``NeurIPS Paper Checklist"},
    \item  {\bf Keep the checklist subsection headings, questions/answers and guidelines below.}
    \item {\bf Do not modify the questions and only use the provided macros for your answers}.
\end{itemize} 

\begin{enumerate}

\item {\bf Claims}
    \item[] Question: Do the main claims made in the abstract and introduction accurately reflect the paper's contributions and scope?
    \item[] Answer: \answerYes{}
    \item[] Justification: The abstract and \Cref{sec:introduction} clearly state the contributions of \name, including the novel factorization architecture, Dual-SVID initialization, and Residual Compensation. Experimental results supporting these claims are provided in \Cref{sec:experiments} and \Cref{sec:analysis}.

\item {\bf Limitations}
    \item[] Question: Does the paper discuss the limitations of the work performed by the authors?
    \item[] Answer: \answerYes{}
    \item[] Justification: We explicitly discuss limitations, such as the lack of compression for the language model head and the computational cost of QAT, in Appendix \ref{appendix:limitation}.

\item {\bf Theory assumptions and proofs}
    \item[] Question: For each theoretical result, does the paper provide the full set of assumptions and a complete (and correct) proof?
    \item[] Answer: \answerYes{}
    \item[] Justification: Mathematical details for the \name forward pass and proofs/claims regarding quantization errors are provided in Appendix \ref{appendix:proofs}.

\item {\bf Experimental result reproducibility}
    \item[] Question: Does the paper fully disclose all the information needed to reproduce the main experimental results of the paper to the extent that it affects the main claims and/or conclusions of the paper (regardless of whether the code and data are provided or not)?
    \item[] Answer: \answerYes{}
    \item[] Justification: We provide full details on model architectures, training settings, and hyperparameters in \Cref{sec:exp_settings} and Appendix \ref{appendix:hyper_parameters} (including \Cref{tab:training_detail}). Specifics on initialization, average bit calculation, and GQA adjustments are in \Cref{sec:dual_svid}, Appendix \ref{appendix:average_bit}, and Appendix \ref{appendix:gqa}.

\item {\bf Open access to data and code}
    \item[] Question: Does the paper provide open access to the data and code, with sufficient instructions to faithfully reproduce the main experimental results, as described in supplemental material?
    \item[] Answer: \answerYes{}
    \item[] Justification: Our code is provided as supplemental material and will be publicly released. We use open datasets (WikiText-2, C4) and standard open-source models (Llama, OPT, etc.).

\item {\bf Experimental setting/details}
    \item[] Question: Does the paper specify all the training and test details (e.g., data splits, hyperparameters, how they were chosen, type of optimizer, etc.) necessary to understand the results?
    \item[] Answer: \answerYes{}
    \item[] Justification: Comprehensive training details, including optimizer settings, learning rate schedules, and GPU configurations, are provided in \Cref{sec:exp_settings} and Appendix \ref{appendix:hyper_parameters}.

\item {\bf Experiment statistical significance}
    \item[] Question: Does the paper report error bars suitably and correctly defined or other appropriate information about the statistical significance of the experiments?
    \item[] Answer: \answerNo{}
    \item[] Justification: Due to the high computational cost of performing QAT on large language models (up to 32B parameters), we report results from single runs, consistent with standard practices in LLM quantization research (see \Cref{tab:main_results}).

\item {\bf Experiments compute resources}
    \item[] Question: For each experiment, does the paper provide sufficient information on the computer resources (type of compute workers, memory, time of execution) needed to reproduce the experiments?
    \item[] Answer: \answerYes{}
    \item[] Justification: We specify the hardware used (NVIDIA A100/H100 GPUs) and the number of GPUs per experiment in \Cref{sec:exp_settings}, \Cref{sec:latency}, and \Cref{tab:training_detail}.

\item {\bf Code of ethics}
    \item[] Question: Does the research conducted in the paper conform, in every respect, with the NeurIPS Code of Ethics \url{https://neurips.cc/public/EthicsGuidelines}?
    \item[] Answer: \answerYes{}
    \item[] Justification: Our work involves compressing public open-source models and uses standard datasets. It does not involve human subjects or sensitive personal data.

\item {\bf Broader impacts}
    \item[] Question: Does the paper discuss both potential positive societal impacts and negative societal impacts of the work performed?
    \item[] Answer: \answerYes{}
    \item[] Justification: We discuss the benefits of efficient LLM deployment and the potential risks of easier access to powerful models in Appendix \ref{sec:societal_impact}.

\item {\bf Safeguards}
    \item[] Question: Does the paper describe safeguards that have been put in place for responsible release of data or models that have a high risk for misuse (e.g., pretrained language models, image generators, or scraped datasets)?
    \item[] Answer: \answerYes{}
    \item[] Justification: We address responsible use and safeguards in our Societal Impact discussion in Appendix \ref{sec:societal_impact}. We do not release new base models but compress existing ones.

\item {\bf Licenses for existing assets}
    \item[] Question: Are the creators or original owners of assets (e.g., code, data, models), used in the paper, properly credited and are the license and terms of use explicitly mentioned and properly respected?
    \item[] Answer: \answerYes{}
    \item[] Justification: We cite all original models (Llama, OPT, etc.) and datasets (WikiText-2, C4) in \Cref{sec:exp_settings} and throughout the text. We respect the licenses of these open-source assets.

\item {\bf New assets}
    \item[] Question: Are new assets introduced in the paper well documented and is the documentation provided alongside the assets?
    \item[] Answer: \answerYes{}
    \item[] Justification: The code for \name (our primary new asset) is documented and included in the supplemental material.

\item {\bf Crowdsourcing and research with human subjects}
    \item[] Question: For crowdsourcing experiments and research with human subjects, does the paper include the full text of instructions given to participants and screenshots, if applicable, as well as details about compensation (if any)? 
    \item[] Answer: \answerNA{}
    \item[] Justification: This research did not involve human subjects or crowdsourcing.

\item {\bf Institutional review board (IRB) approvals or equivalent for research with human subjects}
    \item[] Question: Does the paper describe potential risks incurred by study participants, whether such risks were disclosed to the subjects, and whether Institutional Review Board (IRB) approvals (or an equivalent approval/review based on the requirements of your country or institution) were obtained?
    \item[] Answer: \answerNA{}
    \item[] Justification: This research did not involve human subjects.

\item {\bf Declaration of LLM usage}
    \item[] Question: Does the paper describe the usage of LLMs if it is an important, original, or non-standard component of the core methods in this research? Note that if the LLM is used only for writing, editing, or formatting purposes and does not impact the core methodology, scientific rigorousness, or originality of the research, declaration is not required.
    \item[] Answer: \answerNA{}
    \item[] Justification: LLMs are the subject of our study (we are compressing them), but we did not use LLMs to generate the scientific content or methodology of this paper.

\end{enumerate}

\end{document}